\documentclass[journal,twoside,web]{ieeecolor}
\usepackage{tmi}
\usepackage{cite}
\usepackage{amsmath,amssymb,amsfonts}
\usepackage{algorithmic}
\usepackage{graphicx}
\usepackage{textcomp}

\usepackage{subcaption}
\usepackage{pifont}
\usepackage{booktabs}
\usepackage[export]{adjustbox}
\usepackage{multirow}
\usepackage{makecell}
\usepackage{hyperref} 
\usepackage{color}
\usepackage{multirow}
\usepackage{array}
\usepackage{booktabs} % 优美的表格线条
\usepackage{caption}
\usepackage{xcolor}
\definecolor{myred}{rgb}{0.996,0.578,0.574}
\definecolor{myyellow}{rgb}{0.988,0.961,0.898}
\definecolor{mylightred}{rgb}{0.992,0.887,0.883}
\definecolor{mylightblue}{rgb}{0.686, 0.933, 0.933}  % 浅蓝色
\definecolor{mugreen}{RGB}{0,100,0}
\usepackage[table,xcdraw]{xcolor}
\usepackage{tabularx}
\newcolumntype{Y}{>{\raggedright\arraybackslash}X}
\usepackage[ruled,vlined]{algorithm2e}
\definecolor{lightgreen}{rgb}{0.56, 1.0, 0.56} % 浅绿色
\definecolor{lightcoral}{rgb}{1.0, 0.73, 0.73} % 浅红色
\usepackage{makecell}  % For \makecell
\usepackage{longtable}
\usepackage{afterpage}
\def\BibTeX{{\rm B\kern-.05em{\sc i\kern-.025em b}\kern-.08em
    T\kern-.1667em\lower.7ex\hbox{E}\kern-.125emX}}
\begin{document}
\title{OralAgent: Integrating Reasoning, Tools, and Knowledge for Interactive Dental Image Analysis}
\author{Jing Hao, Siyuan Dai, Yongxin Zhang, Yuci Liang, Jiamin Wu, Jiahao Bao, Yuxuan Fan, \\
Zanting Ye, Yanpeng Sun, Xinyu Zhang, Ming Hu, Liang Zhan, \\
James Kit Hon Tsoi, Linlin Shen, Junjun He, Kuo Feng Hung
\thanks{Corresponding author: Kuo Feng Hung}
\thanks{Jing Hao, Yongxin Zhang, Jiamin Wu, James Kit Hon Tsoi, and Kuo Feng Hung are with the Faculty of Dentistry, the University of Hongkong, Hong Kong SAR, China (e-mail: jinghao@connect.hku.hk, jaminwoo@connect.hku.hk, jkhtsoi@hku.hk, hungkfg@hku.hk).}
\thanks{Siyuan Dai and Liang Zhan are with the Department of Electrical and Computer Engineering, University of Pittsburgh, Pittsburgh, PA, USA. (e-mail: siyuan.dai@pitt.edu, liang.zhan@pitt.edu).}
\thanks{Yuci Liang and Linlin Shen are with Shenzhen University, China (e-mail: yucliang743@gmail.com, llshen@szu.edu.cn).}
\thanks{Jiahao Bao is with Department of Craniomaxillofacial Surgery, Shanghai Ninth People's Hospital, China (e-mail:baojh0123@163.com).}
\thanks{Yuxuan Fan is with the Nanyang technological University, Singapore. (e-mail: orionisfan@gmail.com)}
\thanks{Zanting Ye is with School of Biomedical Engineering, Southern Medical University, China (e-mail: yzt2861252880@gmail.com).}
\thanks{Yanpeng Sun is with Singapore University of Technology and Design, Singapore (yanpeng\_sun@sutd.edu.sg).}
\thanks{Xinyu Zhang is with the University of Auckland, new zealand (xyzhang0717@gmail.com).}
\thanks{Ming Hu and Junjun He are with Shanghai Artificial Intelligence Laboratory , China (e-mail: huming@pjlab.org.cn, hejunjun@sjtu.edu.cn).}
}
 
\maketitle

\begin{abstract}
Dental image analysis plays a pivotal role in supporting accurate diagnosis and treatment planning in oral healthcare. 
Although recent advances have produced dental AI models for specific tasks and individual imaging modalities, their isolated designs limit practical use in real-world clinical workflows.
In this paper, we present OralAgent, the first dental-specialized AI agent that unifies multimodal reasoning, tool‑based decision-making, and knowledge-grounded retrieval within an end-to-end automated framework. 
It integrates 22 visual analysis tools and 368 widely-used classical dental textbooks, enabling autonomous reasoning, planning, tool use, knowledge retrieval, and multi-step workflow execution. 
Furthermore, we introduce OralCorpus, a large‑scale, high‑quality bilingual textual resource containing 134.8M tokens curated for dental retrieval‑augmented generation (RAG). 
To evaluate models’ multidisciplinary dental knowledge, we construct OralQA‑ZH, a Chinese multiple‑choice question benchmark consisting of 798 items across 11 oral subspecialties. 
Extensive experiments demonstrate that OralAgent achieves state‑of‑the‑art performance on the MMOral-Uni, MMOral-OPG, and OralQA‑ZH benchmarks, highlighting its effectiveness, interpretability, and adaptability in real‑world clinical settings. 
The code and models are publicly available at \underline{https://github.com/isjinghao/OralAgent.}
\discretionary{}{}{}\underline{}
\end{abstract}


\section{Introduction}
\label{sec:introduction}
Dentistry has rapidly advanced toward digitalization and semi-automation over the past two decades, largely attributed to its significant dependency on advanced imaging techniques and artificial intelligence (AI). The integration of AI has shown promising potential to improve the accuracy and efficiency of disease detection, diagnosis, and treatment planning~\cite{hao2024semi, hao2024semit, hao2024tmamba, hao2025oraldataset}. Most current AI models are designed for narrowly defined tasks within specific dental imaging modalities, such as anatomical and pathological segmentation on panoramic radiographs~\cite{hao2024semi, hao2024tmamba}, disease detection on intraoral images~\cite{hao2026photography}, landmark identification on cephalometric radiographs~\cite{wu2024cephalometric}, and classification of oral squamous cell carcinoma on histopathological images~\cite{guan2026high}.

\begin{figure}[!t]
\centering
\includegraphics[width=1.0\linewidth]{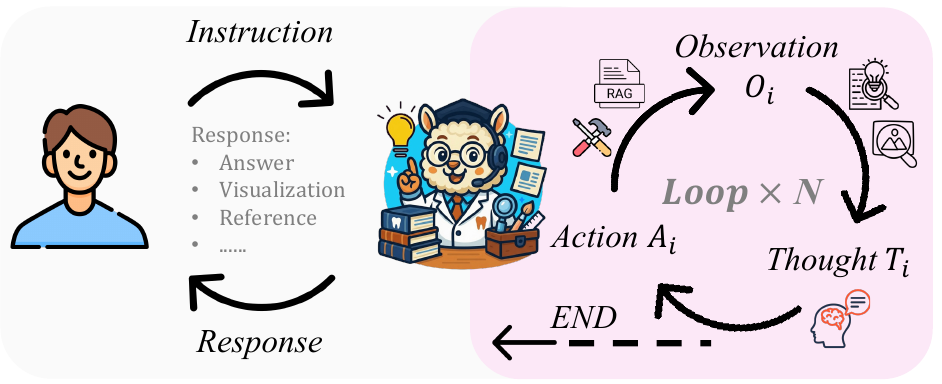}
\caption{The interaction paradigm between the user and our OralAgent. Users submit instructions to the OralAgent, which then begins reasoning based on the user’s intent and initiates an observation–thought–action loop until it produces a final response. OralAgent produces interpretable and reliable responses by generating visual explanations and by providing knowledge reference sources.}
\label{fig:paradigm}
\end{figure}

Although these task-specific models perform well within their respective scopes, their isolated and fragmented design limits their utility in real clinical workflows. In practice, the substantial heterogeneity across dental imaging modalities and the inherent complexity of diagnostic processes require a flexible and versatile AI system capable of handling a wide range of scenarios, rather than a collection of single-purpose models.
Multimodal large language models (MLLM) offer a promising path toward addressing this fragmentation~\cite{sun2024descriptive, hao2024fullanno}. Models such as OralGPT-Omni~\cite{hao2025oralgpt-omni}, trained on over sixty thousand high-quality TRACE-CoT samples, demonstrate broad capabilities in visual question answering (QA) across six dental imaging modalities, while DentalGPT~\cite{cai2025dentalgpt} focuses on interpreting intraoral images and panoramic radiographs. Despite these advances, existing dental MLLMs still face hallucination issues and often fail to systematically evaluate all relevant anatomical structures or integrate findings across different image regions~\cite{liu2025performance1}. Beyond these performance limitations, a more critical concern is their lack of transparency and interpretability, both of which are essential for deployment in real clinical workflows.

To build an AI system with sufficient flexibility and interpretability for use in dental practice, we propose OralAgent, the first dental-specialized AI agent framework for comprehensive dental image analysis. OralAgent dynamically and autonomously performs reasoning, planning, tool invocation, knowledge retrieval, and multi-step workflow execution for dental applications. It combines multimodal reasoning with structured tool-based decision-making and knowledge-grounded retrieval to ensure transparency and reliability. As illustrated in Figure~\ref{fig:paradigm}, OralAgent follows the ReAct loop paradigm~\cite{yao2022react}, decomposing complex user queries into sequential steps involving observation, thought, and action. The system integrates 22 dental-specialized visual tools and 368 widely used classical English and Chinese textbooks covering more than 16 dental disciplines, allowing it to flexibly leverage tools and knowledge sources to address complex queries. Utilizing these dental textbooks, we construct OralCorpus, a large‑scale, high‑quality bilingual textual resource comprising 134.8 million tokens, curated specifically for dental retrieval-augmented generation (RAG). Importantly, OralAgent is highly configurable with respect to its core orchestrator, toolkit, and knowledge base, allowing users to adapt the system to their clinical needs and ensuring strong maintainability and scalability.

At present, there is no text-only dental QA benchmark for systematically evaluating the performance of language models or agent systems. To fill this gap and guide future optimization, we introduce OralQA-ZH, the first benchmark designed to assess mastery of multidisciplinary dental knowledge. 
% All QA pairs are collected and validated by an experienced dentist to ensure clinical accuracy. 
All QA pairs were collected through direct extraction from the included dental textbooks. To ensure their accuracy, a dental professional verified that for each question the provided answer matched the correct option as indicated in the textbook, ensuring that there were no discrepancies or errors between the stated answer and the textbook’s correct choice. Additionally, a senior academic dental professional randomly reviewed 20\% of the pairs, all of which were confirmed to be correct.

We evaluate OralAgent on three benchmarks, MMOral-Uni~\cite{hao2025oralgpt-omni}, MMOral-OPG~\cite{hao2025mmoral}, and OralQA-ZH, to thoroughly assess its capabilities in dental applications. OralAgent achieves new state-of-the-art scores of 57.70 on MMOral-Uni and 61.00 on MMOral-OPG, exceeding the performance of OralGPT Omni by 5.86 points and 15.69 points, respectively. Extensive experiments on OralQA-ZH across various MLLMs and agent systems further show that the RAG module in OralAgent consistently improves performance, demonstrating that retrieved knowledge enhances understanding of multidisciplinary dental topics. We additionally analyze OralAgent's tool use patterns and present case studies illustrating its advantages in dental applications. Finally, ablation studies confirm the effectiveness of its input instruction comprehension module, including intent recognition and modality classification. Our main contributions can be summarized as follows:

\begin{itemize}
    \item We propose OralAgent, the first dental-specialized AI agent framework for dental image analysis that integrates planning, tool invocation, and knowledge retrieval with strong interpretability, reliability, and flexibility. % , without requiring additional training.
    \item We introduce OralCorpus, a large scale, high quality bilingual textual resource containing 134.8 million tokens curated for dental RAG utilization.
    \item We present OralQA-ZH, the first benchmark for evaluating mastery of multidisciplinary dental knowledge, consisting of 798 multiple-choice questions across eleven oral subspecialties.
    \item Extensive experiments show that OralAgent delivers superior performance on MMOral-Uni, MMOral-OPG, and OralQA-ZH, demonstrating high capacity, interpretability, and flexibility for real dental applications.
\end{itemize}

\begin{figure*}[!ht]
\centering
\includegraphics[width=1.0\linewidth]{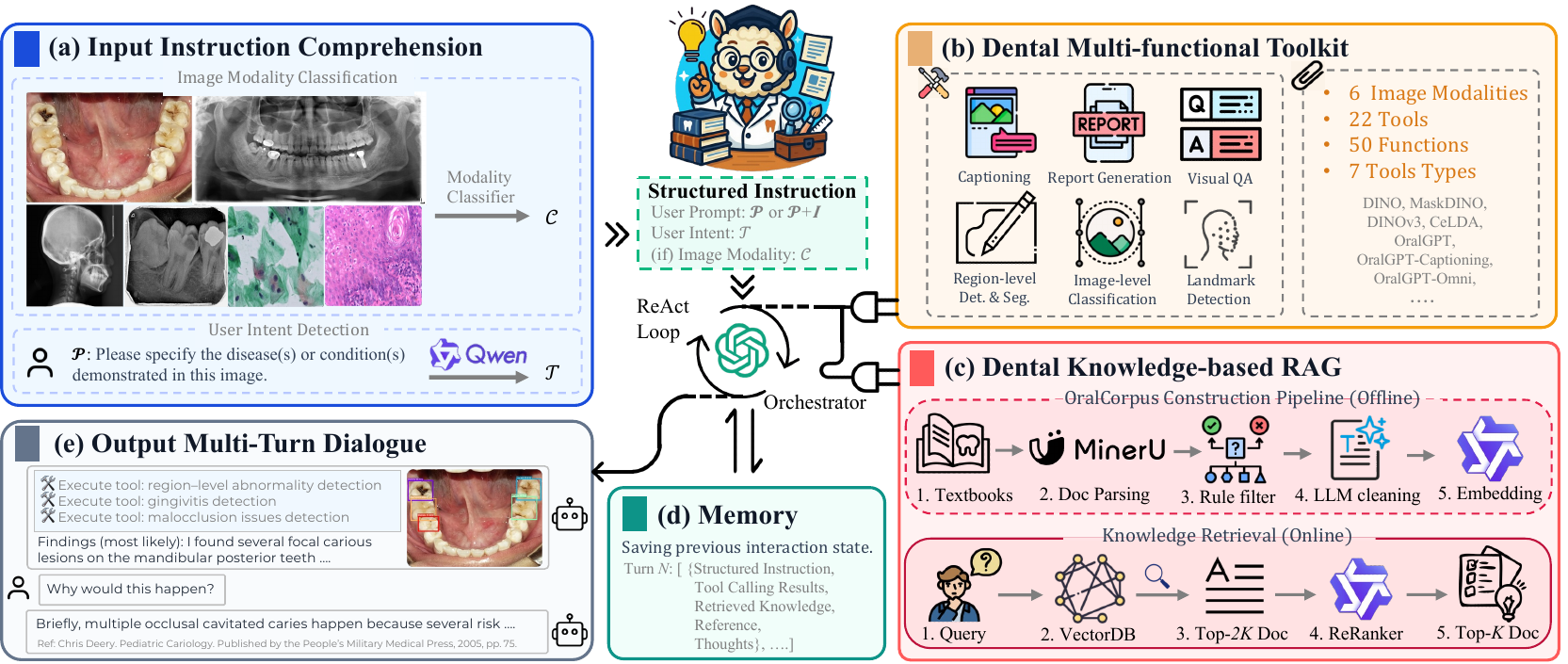}
\caption{Overview of the OralAgent system. (a) The system detects user intent and automatically classifies any uploaded image modality to enable accurate tool selection and invocation. The identified intent is converted into a structured instruction for the core orchestrator. (b) The multi‑functional dental toolkit integrates tools that cover six imaging modalities and more than fifty diagnostic functions. (c) The dental knowledge‑based RAG module integrates 368 textbooks and also supports user‑provided private knowledge bases processed through offline embedding. During online retrieval, it selects the top-k knowledge chunks relevant to the user’s intent together with their sources. (d) The memory module stores the complete interaction history and system state as necessary context for reasoning. (e) OralAgent generates responses that include tool execution details, visualized image analysis, and retrieved knowledge with reference.
}
\label{fig:framework}
\end{figure*}

%--- Section ---%

\section{Related Works}
\label{sec:Related Works}

\noindent
\textbf{Medical Large Language Models.}
Medical MLLMs have advanced the medical field and exhibited immense potential across numerous specialties by providing an interactive natural language interface for clinical tasks. These models fall into two categories, general-purpose and discipline-specific, defined by their capabilities. General-purpose medical MLLMs aim to serve as foundation models that can process diverse medical modalities and support a broad range of tasks. Discipline-specific MLLMs instead target the clinical problems of a single specialty, such as dermatology [64], ophthalmology [24], chest imaging [19], pathology [26], or pediatrics [54]. Within their domains, they often achieve superior diagnostic performance relative to general-purpose medical MLLMs. Despite this progress, most current medical MLLMs generate responses in a single pass. As a result, they struggle with complex reasoning, evidence-based justification, and iterative decision-making that clinicians routinely perform. This limitation undermines reliability and clinical usability and has become a central bottleneck in their development.

\noindent
\textbf{Medical Agent System.}
To advance the capabilities of medical MLLMs, researchers have developed medical agent systems that integrate multiple expert models and retrieval augmented generation (RAG)~\cite{arslan2024rag} tools to enhance flexibility and reliability, leveraging medical expertise to support systematic and rigorous reasoning. Representative systems include MDAgents~\cite{kim2024mdagents}, which uses LLMs to emulate hierarchical diagnostic workflows spanning individual clinicians and collaborative teams. MMedAgent~\cite{li2024mmedagent} incorporates a broad suite of tools to address multimodal medical tasks. MedAgent‑Pro~\cite{wang2025medagentpro} decomposes diagnosis into sequential components, pairing disease‑level standardized plan generation with patient‑level, stepwise reasoning. MedRAG~\cite{zhao2025medrag} retrieves clinical guidelines and privacy‑sensitive electronic health records to improve adherence to standards and reduce the risk of misdiagnosis. MedRAX~\cite{fallahpour2025medrax} focuses on chest X-ray, integrating multiple analysis tools and dynamically orchestrating specialized modules for complex queries, thereby mitigating hallucinations and inconsistencies in reasoning. Despite these advances, current medical agents still underperform in dentistry, primarily due to the scarcity of high‑performance dental tools for visual perception and large-scale corpora for information retrieval. This study addresses these gaps by introducing a dental‑specialized agent system that supports robust and comprehensive dental imaging analysis across multiple modalities and clinical tasks.

\noindent
\textbf{Dental Image Analysis.}
AI in dentistry has advanced markedly in recent years. It now supports tasks ranging from tooth segmentation and disease detection to treatment planning and digital crown design, demonstrating substantial promise for both clinical and laboratory workflows. Early work focused on neural networks designed for single tasks such as tooth segmentation, plaque detection, tumor prediction, and so on. More recently, efforts have shifted toward dentistry-specific MLLMs. OralGPT~\cite{hao2025mmoral}, trained on a large-scale instruction dataset, is an dental MLLM tailored to panoramic X-ray analysis. OralGPT-Plus~\cite{fan2026oralgpt-plus} is  an agentic MLLM designed to perform iterative and symmetry-aware diagnostic reasoning for panoramic dental radiograph analysis. DentVLM~\cite{meng2025dentvlm} supports basic oral disease diagnosis across three common imaging modalities but lacks the ability to provide detailed explanations. DentalGPT~\cite{cai2025dentalgpt} captures fine-grained visual cues and enables more reliable disease-related reasoning. OralGPT-Omni~\cite{hao2025oralgpt-omni} broadens analysis to seven dental imaging modalities and also supports text-only interactions. OPGAgent~\cite{yu2026opgagent} orchestrates specialized perception modules through a consensus mechanism to deliver more accurate and auditable interpretations of orthopantomograms. Despite this progress, the field still lacks a comprehensive, versatile dental agent capable of performing multiple complex tasks across diverse imaging modalities while engaging in accurate, explainable, reference-grounded, and reliable interactions with users.

%--- Section ---%
\section{Methodology}
\label{sec:Methodology}
\subsection{Overview of OralAgent}
OralAgent is an intelligent and flexible agentic system capable of dynamically and autonomously performing reasoning, planning, tool invocation, knowledge retrieval, and execution of workflows with multiple steps for dental applications. Compared to previous task-specific dental AI models, OralAgent intergrates multimodal reasoning with structured tool-based decision-making and knowledge-based RAG, thereby emphasizing interpretability and reliability by incorporating numerous off-the-shelf dental tools and authoritative dental textbooks as well as clinical guidelines. Specifically, OralAgent comprises five components: a core orchestrator, user instruction comprehension, dental toolkit integration, dental RAG integration, and memory saver. The system follows the ReAct loop paradigm~\cite{yao2022react}, which decomposes complex user queries into sequential analytical steps that include observation, thought, and action. It processes a user prompt iteratively by first inferring user intent, then allowing the core orchestrator to reason and plan in order to produce structured tool calls or RAG actions, and finally synthesizing tool outputs and retrieved knowledge to generate the next observation in the cycle. This loop continues until the system produces a final response or requests additional input from the user. An overview and the processing pipeline of OralAgent are presented in Figure~\ref{fig:framework} and Algorithm~\ref{tab:algorithm}.
To mitigate erroneous responses caused by incorrect outcomes from the tool calls, we employ prompt engineering~\cite{wang2025prompt} to explicitly instruct the core orchestrator to rely on its own parametric knowledge when deciding whether to accept tool outputs. For example, we use the prompt: ``\textit{Critically evaluate tool outputs, challenge them when warranted, reconcile inconsistencies, and explain your confidence.}'' This guidance encourages independent verification, reduces the propagation of tool-induced errors, and improves overall model robustness.

\subsection{User Instruction Comprehension}
Recognizing user intent from dialogue utterances is essential for delivering satisfactory responses, as it enables automatic, accurate, and flexible interpretation of user instructions~\cite{rodriguez2024intentgpt}. We implement an intent recognition module that identifies potential intents across diverse and dynamic instructions. This module supports the orchestrator in reasoning and planning the next action at each dialogue turn, and it promotes efficient use of tool calls and knowledge-base retrieval. Based on brainstorming with professional dentists, we define nine intent categories that span multiple user groups, including laypersons, junior dentists, clinicians, and oral health researchers, when they use the dental specialized agent system. The categories include visual feature description, anomaly diagnosis, report generation, treatment planning, prognosis prediction, subtype grading and classification, education, scientific research, and out-of-scope (OOS) queries. We include the OOS category to enable the system to reject queries unrelated to dentistry and to reduce the risk of misinformation outside the field. For intent recognition, we use the Qwen3-0.6B model~\cite{yang2025qwen3} within an in-context learning framework, and design the prompt carefully to guide the model for the intent detection. At each dialogue turn, the model processes the prompt together with the user utterance, infers the underlying intent, and outputs one or more labels from the predefined dentistry-related taxonomy.
In addition to processing textual instructions, OralAgent analyzes multiple dental imaging modalities, which are indispensable sources of information in dentistry. To address the heterogeneity and complexity across modalities that support distinct tasks, we perform modality classification as a preprocessing step to enhance automation and adaptability of the system. The resulting modality labels enable the orchestrator to interpret user instructions more accurately, generate precise plans, and invoke the appropriate modality-specific visual tools during reasoning. We finetune BioMedCLIP~\cite{zhang2023biomedclip}, a biomedical foundation model, for dental imaging modality classification using 78,221 dental-related images spanning six modalities. Integrating the multimodal interpretation of the user's prompt, which includes the textual instruction and the visual modality, with the original instruction, we pass this structured  instruction to the orchestrator to plan the subsequent problem solving process. This integration enables OralAgent to capture user intent more effectively and provides the orchestrator with richer semantic context for reasoning, planning, tool invocation, and knowledge retrieval, thereby producing more accurate and reliable responses.

\begin{algorithm}[t]
% \small
\caption{OralAgent ReAct Framework}
\label{tab:algorithm}
\SetKwInOut{Input}{Input}
\SetKwInOut{Output}{Output}

\Input{
$Q$: User query\\
$I$: Set of dental images (if any)\\
$T$: Available dental AI tools\\
$K$: Available dental knowledge bases\\
$M$: Memory buffer\\
$t_{\max}$: Maximum allowed time
}

\Output{$R$: Final response to query}

\textbf{Initialize:}\\
$t_{\text{start}} \gets \mathrm{GetCurrentTime}()$\\
${\mathcal{T}, \mathcal{C}} \gets \mathrm{InstructionComprehension}(Q,I)$\\

$\mathit{state} \gets \mathrm{Observe}(Q, I, \mathcal{T}, \mathcal{C}, M)$\\

\While{$\mathrm{GetCurrentTime}() - t_{\text{start}} < t_{\max}$}{
$\mathit{thoughts} \gets \mathrm{Reason}(\mathit{state}, M)$\;
\If{$\mathrm{RequiresUserInput}(\mathit{thoughts})$}{
\Return $\mathrm{GenerateUserPrompt}(\mathit{thoughts}, M)$\;
}
\If{$\mathrm{CanGenerateResponse}(\mathit{thoughts})$}{
\Return $\mathrm{GenerateResponse}(\mathit{thoughts}, M)$\;
}
$\mathit{tools} \gets \mathrm{SelectTools}(\mathit{thoughts}, T, M)$\;
$\mathit{results} \gets \mathrm{ExecuteParallel}(\mathit{tools}, \mathit{state})$\;
$\mathit{knowledge} \gets \mathrm{RAG}(\mathit{state},K,M)$\;

$M \gets M \cup \{(\mathit{thoughts}, \mathit{tools}, \mathit{results}, \mathit{knowledge})\}$\;
$\mathit{state} \gets \mathrm{Observe}(\mathit{state}, \mathit{results}, \mathit{knowledge}, M)$\;
}
\Return $\mathrm{GenerateTimeoutResponse}(\mathit{state}, M)$
\end{algorithm}
\begin{table*}[t]
\centering
\footnotesize
\setlength{\tabcolsep}{5pt}
\renewcommand{\arraystretch}{1.2}
\rowcolors{2}{gray!6}{white}
\caption{OralAgent integrates 22 tools across six dental imaging modalities, spanning six tasks, including classification, object detection, segmentation, keypoint detection, report generation, and visual QA. For reference, the table lists each tool’s supported functions, source, and performance.}
\label{tab:toolkit}
\begin{tabularx}{\textwidth}{
    >{\centering\arraybackslash}p{0.045\textwidth}
    p{0.165\textwidth}
    p{0.12\textwidth}
    p{0.10\textwidth}
    Y
    >{\centering\arraybackslash}p{0.06\textwidth}
    >{\centering\arraybackslash}p{0.10\textwidth}
}
\toprule
\textbf{No.} & \textbf{Modality} & \textbf{Task} & \textbf{Tool} & \textbf{Supportive functions} & \textbf{Source} & \textbf{Performance} \\
\midrule
T01 & Intraoral image & Classification & DINOv3 & Calculus, Caries, Gingivitis, Ulcer, Tooth discoloration, Defective dentition, Cancer (malignant lesion) & in-house & Acc=99.9\% \\
% 2 & Intraoral image & Segmentation & MaskDINO & Dental Plaque & mAP50=46.1 \\
T02 & Intraoral image & Detection & DINO & Abrasion, Filling, Crown, Caries & in-house & mAP50=92.2 \\
T03 & Intraoral image & Detection & DINO & Gingivitis &in-house & mAP50=78.4 \\
T04 & Intraoral image & Detection & DINO & Fenestration and Dehiscence &in-house & mAP50=71.9 \\
T05 & Intraoral image & Detection & DINO & Tooth torsion, Tooth emergence, Invisible orthodontic attachment, Fixed orthodontic device, case-fixed orthodontic appliances, Tooth misalignment, Mandibular retrusion, Orthodontic brace &in-house & mAP50=73.9 \\
T06 & Intraoral image & Detection & DINO & 1st Molar, 1st Premolar, 2nd Molar, 2nd Premolar, Canine, Central Incisor, Lateral Incisor &in-house & mAP50=96.7 \\
T07 & Periapical radiograph & Segmentation & MaskDINO & Crown, Caries, Root canal treatment, Restoration, Bone loss, Periodontitis, Apical periodontitis & ~\cite{hao2025mmoral} & mAP50=99.0 \\
T08 & Periapical radiograph & Classification & ViT & Pulpitis, Impacted Tooth, Apical periodontitis, Bone loss, Mixed dentition, Caries, Periodontitis & ~\cite{hao2025mmoral} & Acc=98.7\% \\
T09 & Cephalometric radiograph & Keypoint detection & CeLDA & 29 cephalometric landmarks &in-house & SDR@2=67.06\%  \\
T10 & Histopathology & Classification & DINOv3 & Oral submucous fibrosis (OSMF) and oral squamous cell carcinoma (OSCC) &in-house & Acc=99.3\% \\
T11 & Histopathology & Classification & DINOv3 & OSCC, Leukoplakia w/wo dysplasia &in-house & Acc=88.7\% \\
T12 & Histopathology & Segmentation & MaskDINO & OSCC &in-house & mAP50=99.9 \\
T13 & Cytopathology & Segmentation & MaskDINO & Abnormal epithelial nucleus, healthy epithelial nucleus, out-of-focus nucleus, blood cell nucleus, reactive cell nucleus, dividing nucleus &in-house & mAP50=94.2 \\
T14 & Cytopathology & Segmentation & MaskDINO & Cellular anomaly grading &in-house & mAP50=62.5 \\
T15 & Multi-modality & Visual QA & OralGPT-Omni & Answer open-ended, dentistry-specific questions. &~\cite{hao2025oralgpt-omni} & SOTA on MMOral-Uni \\
T16 & Multi-modality & Visual Description & OralGPT-Captioning & Describe visual features showed in the image &~\cite{hao2025oralgpt-omni} & - \\
T17 & Panoramic radiograph & Detection & DINO & 1 to 32 tooth numbering following the FDI tooth numbering system & ~\cite{hao2024semit} & mAP50=97.1 \\
T18 & Panoramic radiograph & Detection & DINO & Periapical lesion (Granuloma, Cyst, and Abscess) & ~\cite{hao2025mmoral} & mAP50=98.8 \\
T19 & Panoramic radiograph & Segmentation & MaskDINO & Bone loss & ~\cite{hao2025mmoral} & mAP50=83.5 \\
T20 & Panoramic radiograph & Segmentation & MaskDINO & Caries, Periapical lesions, Impacted tooth, Filling, Crown, Malaligned, Edentulous site, Retained root, Root canal treatment, Mandibular canal, Maxillary sinus & ~\cite{hao2025mmoral} & mAP50=69.6 \\
T21 & Panoramic radiograph & Segmentation & MaskDINO & Mandibular canal, Maxillary sinus & ~\cite{hao2025mmoral} & mAP50=88.5 \\
T22 & Panoramic radiograph & Report Generation & OralGPT & Generate a comprehensive report for panoramic x-ray analysis. & ~\cite{hao2025mmoral} & SOTA on MMOral-OPG \\

\bottomrule
\end{tabularx}
\end{table*}

\subsection{Dental Tool Integration}
OralAgent integrates a diverse suite of dental-specialized visual models to support comprehensive imaging analysis. All expert models are organized into a multifunctional toolbox that can be autonomously selected and invoked according to user requirements. The tools accessible within each imaging modality and their supported functions are summarized in Table~\ref{tab:toolkit}. Specifically, OralAgent unifies 22 visual expert models across six imaging modalities, including intraoral images, panoramic radiographs, periapical radiographs, cephalometric radiographs, histopathology, and cytopathology. These experts cover a broad range of downstream tasks, such as visual captioning, visual question answering, report generation, abnormality classification, disease detection, landmark identification, etc.
The dental-specialized toolkit comprises two sources, including existing open-source models and newly developed in-house models. We carefully selected ten powerful open-source experts based on task-specific results, including SemiT-SAM~\cite{hao2024semit}, OralGPT~\cite{hao2025mmoral}, OralGPT-Omni~\cite{hao2025oralgpt-omni}, OralGPT-Captioning~\cite{hao2025oralgpt-omni}, and six dental visual expert models~\cite{hao2025mmoral}. SemiT-SAM~\cite{hao2024semit} performs teeth instance segmentation by delineating 32 teeth and assigning tooth numbering IDs according to the FDI World Dental Federation notation in panoramic and periapical radiographs. OralGPT~\cite{hao2025mmoral} is a multimodal large language model specialized for report generation on panoramic radiographs. OralGPT-Omni~\cite{hao2025oralgpt-omni} is a versatile MLLM designed for visual question answering across the six dental imaging modalities. The remaining 6 expert models~\cite{hao2025mmoral} are designed to detect a range of conditions and diseases on panoramic radiographs. These open-source experts demonstrate state-of-the-art (SOTA) performance on their respective tasks and provide accurate and dependable evidence for dental image analysis. 
To further enhance OralAgent for complex real-world scenarios, we developed 12 additional experts by finetuning high-capacity models on multiple publicly available datasets. For image-level classification, we chose DINOv3~\cite{simeoni2025dinov3} as the base model and finetuned four experts. For region-level detection and segmentation, we used DINO~\cite{zhang2022dino} and MaskDINO~\cite{li2023maskdino} as base models due to their strong performance on natural images and finetuned five experts. We also fine-tuned CeLDA~\cite{wu2024cephalometric} on a public dataset to enable landmark detection on cephalometric radiographs. 
During model development, each expert was evaluated on a validation set constructed via an 8:2 random split of the original training set, so as to mitigate the risk of test data leakage in the benchmark. The results, summarized in Table~\ref{tab:toolkit}, confirm the effectiveness of each expert. Collectively, the experts function as a specialist cohort, demonstrating strong proficiency in targeted tasks across diverse dental imaging modalities. 
% Notably, the dental conditions or diseases detected by different experts partially overlap. This mirrors clinical practice, where multiple specialists provide convergent assessments of the same case, thereby improving the accuracy and reliability of the final outputs.

\begin{figure}[!ht]
\centering
\includegraphics[width=1.0\linewidth]{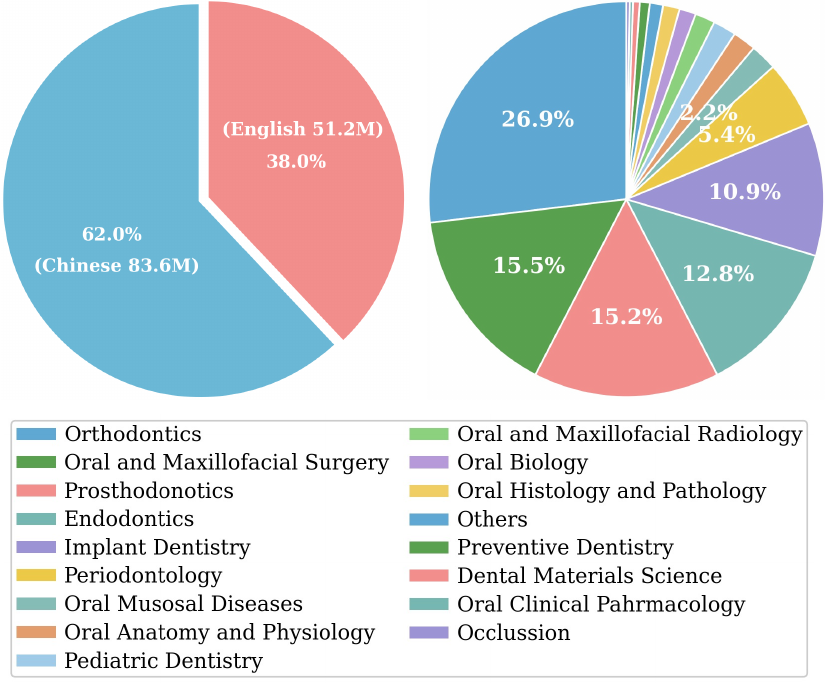}
\caption{Distribution of the proposed OralCorpus. Left: language composition; Right: dental specialty composition}
\label{fig:OralCorpus}
\end{figure}

\subsection{Dental RAG Integration}
% Dentistry is a specialized medical field that relies on a precise terminology system and numerous established truths, such as specific disease symptoms or drug side effects. 
Similar to other medical fields, dentistry relies on precise terminology and well-established knowledge, such as the characteristic symptoms of specific diseases and the potential side effects of medications used in treatment. In this domain, the ability to perform complex reasoning while generating accurate and credible responses backed by verifiable sources, is essential for dental applications. Retrieval-augmented generation (RAG)~\cite{arslan2024rag} is a technique that answers user queries using a specific and private knowledge base without requiring further training of the model. To facilitate digital dentistry, we construct OralCorpus, a large-scale, high-quality, bilingual textual resource with 134.8M tokens curated for dental RAG utilization. 
OralCorpus is built from 331 Chinese dental textbooks and 37 English dental textbooks, followed by a rigorous four-step offline data cleaning pipeline described in Figure~\ref{fig:framework}(c). First, we use MinerU 2.5~\cite{wang2024mineru} to perform document parsing and optical character recognition (OCR) on all textbooks. This converts unstructured books into structured data that include informative titles, paragraphs, tables, page numbers, and non-content elements such as page headers and footers. Second, we apply a rule-based post-processing strategy to refine the parsed results: we filter out dental-unrelated material, including headers and footers; restore sentence integrity by merging fragments split by layout; and assign a page number to every paragraph to enable precise citation in the downstream RAG process. We retain only those pages that contain page numbers because unnumbered pages are typically metadata unrelated to dental knowledge. 
Subsequently, we use an LLM to further enhance corpus quality in three ways. First, we assess each paragraph to determine whether it constitutes dental-domain knowledge suitable for inclusion in the knowledge base. Second, we remove routine noise from the parsed text, such as figure and table references (e.g., \textit{“as shown in Figure ...”}). Third, we translate content into Chinese or English as needed. We carefully designed bilingual prompts and instructed GPT-5-nano to perform this cleaning. Finally, we embed all paragraphs using the Qwen3-Embedding-8B model~\cite{zhang2025qwen3_embed_rerank}, yielding 1.4M chunks in both English (51.2M tokens) and Chinese (83.6M tokens). We implement the dental RAG branch using the LangChain framework. The system indexes the entire OralCorpus as vector embeddings and retrieves the top 2*K candidates based on cosine similarity for each user query. The Qwen3-Reranker-8B~\cite{zhang2025qwen3_embed_rerank} model then reranks these candidates and returns the top-K knowledge items, along with their source book titles and page numbers. These results are passed to the OralAgent orchestrator for subsequent reasoning and response generation.

\subsection{Flexible Modularity and Deployment}
OralAgent is highly flexible, with customization along three dimensions: the core orchestrator, the tool configuration, and the dental knowledge base preparation. The core orchestrator, which serves as the reasoning engine, can be any off-the-shelf LLM or MLLM with structured tool calling capabilities, ranging from open source to proprietary. By default, OralAgent uses GPT-5-nano as the core orchestrator due to its low cost and strong reasoning capabilities. Regarding tool configuration, tools can be added, removed, or replaced for dental-specialized downstream tasks without affecting other functionality. Integration of a new tool is plug-and-play and does not require any post training; users simply specify the tool’s input and output formats, name, and capability description, which enables flexible addition and upgrading of specialized and customized dental visual tools. The knowledge base preparation also supports local, private, and confidential repositories through an offline document parsing pipeline. Users only need to provide the path to their personal knowledge base, with support for \textit{.txt}, \textit{.doc}, \textit{.md}, and \textit{.pdf} files, allowing local integration into OralAgent while preserving privacy and confidentiality. 
Meanwhile, OralAgent supports deployments from local installations to cloud-based solutions, thereby addressing diverse healthcare privacy requirements. It provides two modes of invocation, namely HTTP requests and an interactive interface. HTTP requests are suitable for large-scale batch processing, while the interface, implemented with Gradio, facilitates user-friendly deployment in real-world settings. The interface allows uploading dental images across six modalities and maintains an interactive chat session for natural multi-turn interactions. It further provides transparency into tool execution by tracking and displaying intermediate outputs and reporting the sources of retrieved dental knowledge, for example, book title and page number. This end-to-end implementation enables rapid integration of OralAgent into existing clinical workflows.

\begin{table}[!t]
\centering
\caption{Overview of Category Distribution in the OralQA-ZH Benchmark.}
\label{tab:OralQA-ZH}
\begin{tabular}{l l c}
\toprule
\textbf{Category} & \textbf{Abbrev.} & \textbf{Count} \\
\midrule
Endodontics & Endo & 133 \\
Periodontology & Perio & 89 \\
Oral and Maxillofacial Surgery & OMFS & 180 \\
Prosthodontics & Prosth & 168 \\
Orthodontics & Ortho & 24 \\
Oral Mucosal Diseases & OMD & 54 \\
Pediatric Dentistry & PedDent & 28 \\
Oral and Maxillofacial Radiology & OMFR & 12 \\
Preventive Dentistry & PrevDent & 41 \\
Oral Epidemiology & OralEpi & 29 \\
Oral and Maxillofacial Pathology & OMFP & 40 \\
\midrule
\textbf{Overall} & -- & 798 \\
\bottomrule
\end{tabular}
\end{table}

\begin{figure*}[!ht]
\centering
\includegraphics[width=1.0\linewidth]{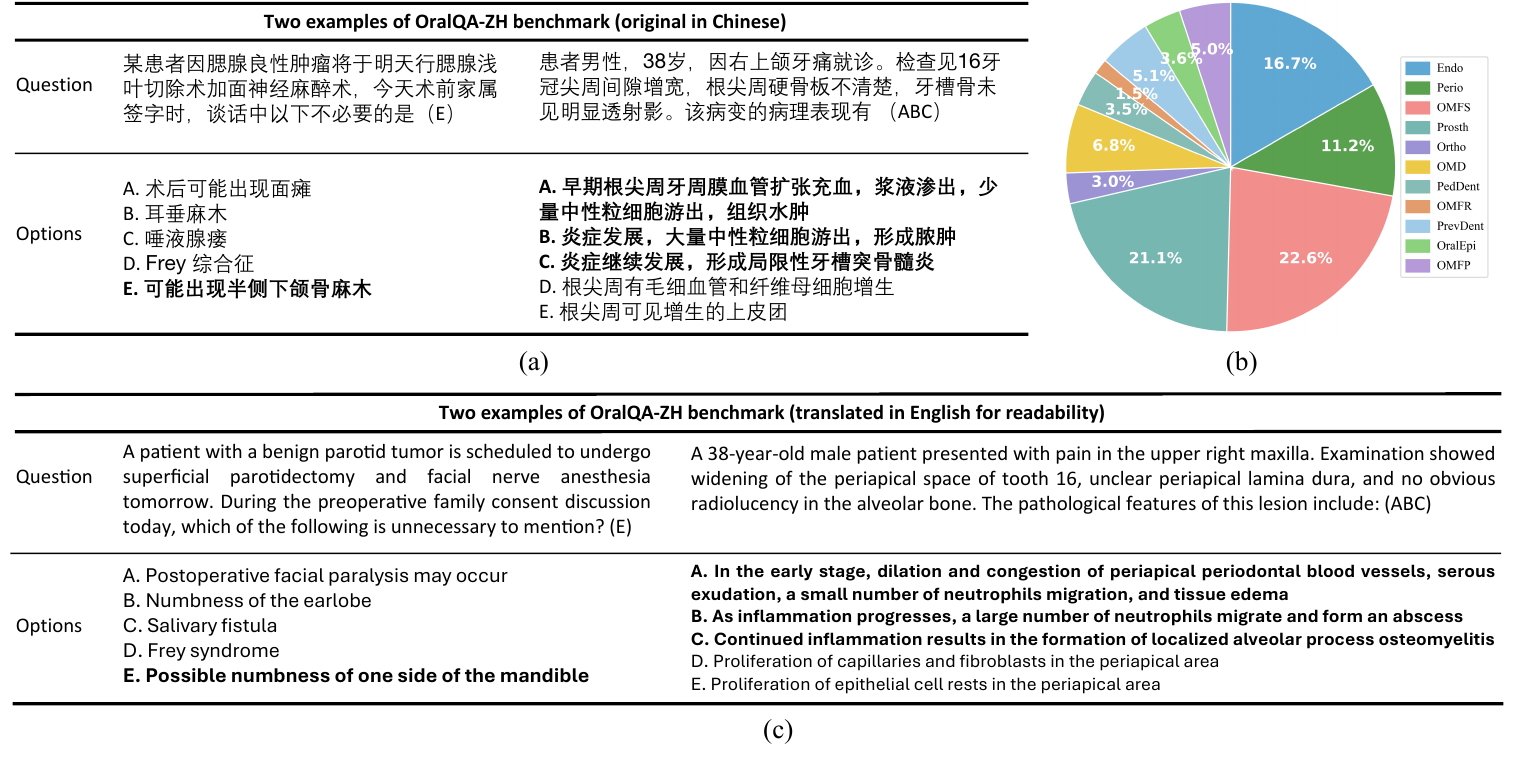}
\caption{Examples and Distribution of the Proposed OralQA‑ZH Benchmark. (a) Two example QA pairs from the OralQA‑ZH Benchmark. All QA items are in multiple‑choice format, and each question may have more than one correct option. (b) Distribution of knowledge categories in the dental domain. (c) English translations of these two Chinses QA pairs for readability.
}
\label{fig:OralQA}
\end{figure*}

%--- Section ---%
\section{OralQA-ZH Benchmark}
\label{sec:OralQA}
\noindent
\textbf{Data Collection.}
We collected questions and corresponding answer candidates from the Mainland China National Dental Physician Licensing Examination. The questions include requests for specific factual knowledge as well as clinical vignettes that describe patient conditions. The tests cover common oral diseases across age groups, related comorbidities, and key diagnostic reasoning. They include etiology, pathogenesis, clinical manifestations, auxiliary examinations, diagnosis and differential diagnosis, and treatment principles. The benchmark reflects real and complex clinical scenarios and therefore requires broad interdisciplinary knowledge to judge the correctness of choices. This design allows us to assess model performance in complex and authentic clinical settings. All questions come from actual and practice examinations that are freely accessible online.

\noindent
\textbf{Data Analysis.}
The OralQA-ZH benchmark is written in simplified Chinese and contains 798 multiple-choice questions across 11 oral subspecialties, which are summarized in Table~\ref{tab:OralQA-ZH} and Figure~\ref{fig:OralQA}. OralQA-ZH uses a closed-ended multiple-choice format in which each question provides several answer options and one or more of them may be correct. For evaluation, we use accuracy as the metric. A model’s predicted answers must match the ground truth exactly to be considered correct. We report evaluation results for each subspecialty as well as the overall performance.

%--- Section ---%
\section{Experiments and Discussion}
\label{sec:Experiments}
\subsection{Implementations}
OralAgent is a ReAct-based~\cite{yao2022react} agentic system that integrates multiple open‑source LLMs and MLLMs. We selected the Qwen3‑0.6B~\cite{yang2025qwen3} for user intent recognition because the task has low complexity and requires high efficiency. GPT‑5‑nano~\cite{gpt5} is used as the core orchestrator due to its strong instruction‑following ability and effective tool‑use capabilities.
OralAgent performs tool execution through structured JSON schema calls, where the core orchestrator formulates  precise JSON-formatted requests that include required arguments such as timestamps, IDs, tool names, and arguments of tools.
For the dental RAG module, the Qwen3‑Embedding‑8B~\cite{zhang2025qwen3_embed_rerank} and the Qwen3‑Reranker‑8B~\cite{zhang2025qwen3_embed_rerank} are used for similarity computation and reranking. OralAgent is built on the LangChain and LangGraph frameworks, and it employs the default \textit{MemorySaver} module for memory management. 
OralAgent is deployed on a single NVIDIA A100 80G GPU, and it integrates 22 tools and a corpus of 368 dental textbooks. In addition, it offers both HTTP endpoints and an interactive interface implemented with Gradio to support flexible user interaction. All code, tools, and knowledge bases are publicly available in our \href{https://github.com/isjinghao/OralAgent}{GitHub repository}.

\begin{figure}[!ht]
\centering
\includegraphics[width=1.0\linewidth]{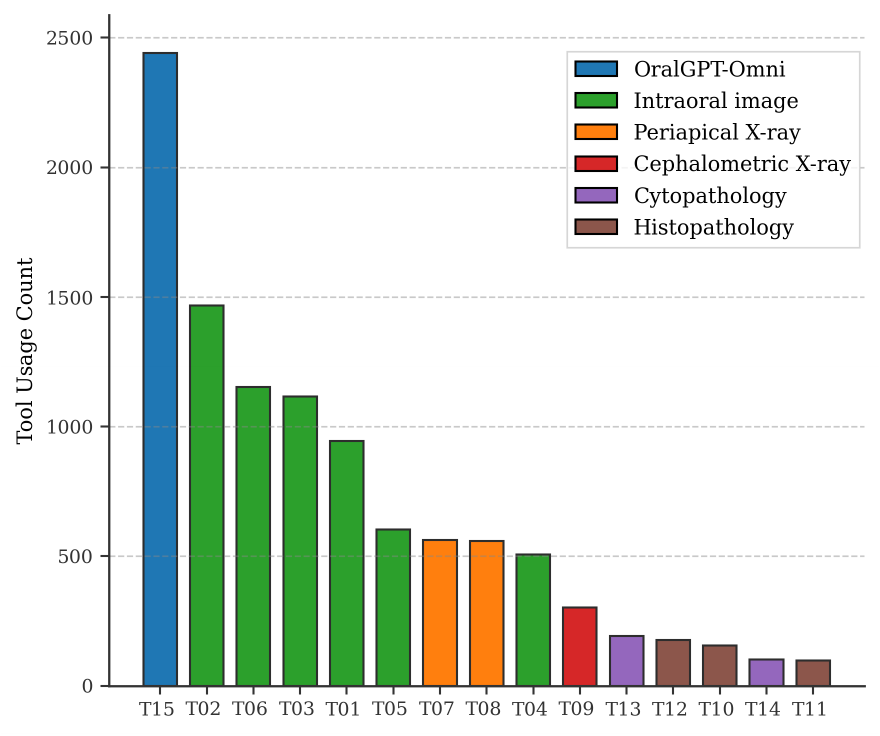}
\caption{Tool use distribution of OralAgent on the MMOral‑Uni benchmark, where the system makes an average of 3.67 tool calls per case.}
\label{fig:tool_usage}
\end{figure}

\begin{table*}[h!]
\centering
\caption{Results on the MMOral-Uni for various existing MLLMs across five dental imaging modalities. The abbreviations are defined as follows: II - Intraoral Image, PA - Periapical Radiograph, CE - Cephalometric Radiograph, PI - Pathological Image, TP - Treatment Planning, IV - Intraoral Video. The best-performing model in each category is highlighted \textbf{in-bold}, while the second-best is \underline{underlined}. }
\vspace{-3pt}
\label{tab:mmoral-uni}
\small % 缩小整体字体以减少宽度
\renewcommand{\arraystretch}{1.0}
\setlength{\tabcolsep}{4pt} % 减小列间距
% \begin{tabular}{
%     >{\centering\arraybackslash}m{3cm}  % Model
%     >{\centering\arraybackslash}m{1.0cm}  % II_1
%     >{\centering\arraybackslash}m{1.0cm}  % II_2
%     >{\centering\arraybackslash}m{1.0cm}  % II_3
%     >{\centering\arraybackslash}m{1.3cm}  % PA
%     >{\centering\arraybackslash}m{1.3cm}  % CE
%     >{\centering\arraybackslash}m{1.3cm}  % HP
%     >{\centering\arraybackslash}m{1.3cm}  % TP
%     >{\centering\arraybackslash}m{1.3cm}  % TE
%     >{\centering\arraybackslash}m{1.3cm}  % IV
%     >{\centering\arraybackslash}m{1.5cm}  % Overall
% }
\begin{tabular}{
    >{\raggedright\arraybackslash}p{4.3cm}  % 第一列 Model
    *{9}{>{\centering\arraybackslash}p{1.15cm}} % 其余九列等宽
}
\toprule
\multirow{2}{*}{Model} 
& \multicolumn{3}{c}{II} 
& \multirow{2}{*}{PA} 
& \multirow{2}{*}{CE} 
& \multirow{2}{*}{PI} 
& \multirow{2}{*}{TP}
& \multirow{2}{*}{IV}
& \multirow{2}{*}{Overall} \\
\cmidrule(lr){2-4}
& Loc & Dx-I & Dx-R &  &  &  &  &  &  \\
% \midrule
\hline
\rowcolor{mylightblue} \multicolumn{10}{l}{\textit{Proprietary MLLMs}} \\ 
\hline
GPT-5~\cite{gpt5}  & 44.60 & 45.24 & 25.16 & 31.43 & {41.27} & {40.52} & \textbf{80.67} & 56.00 & 36.42 \\
o3~\cite{gpto3} & {45.00} & {53.49} & {28.19} & 37.48 & 28.60 & 37.52 & 75.33 & 50.00 & 38.70 \\
Grok-4~\cite{grok4}	& 11.80 & 35.71 & 20.27 & 30.89 & 22.60 & 34.73 & 72.67 & 23.00 & 28.47 \\
Doubao-1.5-vision-pro-32k~\cite{doubao} & 26.90 & 43.33 & 16.97 & {40.48} & 18.20 & 28.49 & 68.00 & 45.00 & 30.67 \\
Gemini-2.5-flash~\cite{team2023gemini} & 31.00 & 51.10 & 19.70 & 37.01 & 36.30 & 35.09 & 73.33 & 46.00 & 35.72 \\
GLM-4.5v~\cite{v2507glm}	& 21.70 & 45.10 & 14.37 & 38.37 & 31.00 & 27.60 & 70.00 & 33.00 & 31.05 \\
\hline
\rowcolor{mylightblue} \multicolumn{10}{l}{\textit{Open-Source MLLMs}} \\ 
\hline
Qwen2.5-VL-7B~\cite{bai2025qwen25vl} & 11.20 & 25.40 & 23.84 & 19.72 & 13.70 & 30.73 & 36.67 & 12.00 & 22.88 \\
Qwen3-VL-8B-Instruct~\cite{yang2025qwen3} & 14.30 & 27.17 & 16.95 & 33.54 & 4.97 & 30.68 & 54.00 & 18.00 & 23.45 \\
Qwen3-VL-235B-A22B~\cite{yang2025qwen3} & 18.70 & 40.00 & 12.53 & 33.62 & 26.70 & 28.75 & 56.00 & 23.00 & 27.83 \\
GLM-4.1V-9B-Thinking~\cite{v2507glm}  & 15.50 & 37.78 & 18.14 & 37.57 & 13.53 & 27.13 & 62.00 & 43.00 & 27.86 \\
InternVL3.5-8B~\cite{wang2025internvl3} & 10.10 & 30.85 & 11.02 & 32.95 & 14.80 & 28.67 & 64.67 & 14.00 & 23.39 \\
LLaVA-v1.6-Mistral-7B~\cite{llava16} & 4.80 & 20.57 & 14.03 & 11.61 & 2.50 & 12.06 & 34.67 & 18.00 & 13.54 \\
LLaVA-OneVision~\cite{li2024llavaonevision} & 7.50 & 23.13 & 17.59 & 14.34 & 0.00 & 11.49 & 40.67 & 5.00 & 15.39 \\
MiMo-VL-7B~\cite{xiaomi2025mimo} & 25.40 & 36.28 & 22.51 & 34.62 & 18.70 & 31.91 & 62.00 & 43.00 & 29.62 \\
Phi-4-multimodal-instruct~\cite{abdin2024phi} & 3.70 & 18.32 & 7.86 & 18.31 & 1.67 & 9.63 & 34.67 & 7.00 & 12.12 \\
Mistral-Small-3.1-24B~\cite{Mistral} & 16.30 & 28.14 & 18.77 & 29.15 & 11.50 & 27.02 & 62.00 & 28.00 & 23.69 \\
R-4B~\cite{yang2025r4b} & 3.80 & 32.29 & 19.16 & 28.96 & 16.60 & 27.89 & 50.67 & 19.00 & 24.94 \\
Ovis2.5-9B~\cite{lu2025ovis2} & 15.40 & 37.47 & 19.76 & 36.70 & 18.13 & 34.33 & 66.00 & {61.00} & 29.60 \\
\hline
\rowcolor{mylightblue} \multicolumn{10}{l}{\textit{Medical MLLMs}} \\ 
\hline
LLaVA-Med~\cite{li2023llavamed} & 5.40 & 3.99 & 5.81 & 25.31 & 1.50 & 16.08 & 28.67 & 14.00 & 10.16 \\
HuatuoGPT-Vision-7B~\cite{chen2024huatuogpt} & 11.90 & 30.19 & 23.10 & 25.79 & 19.57 & 28.02 & 41.33 & 19.00 & 25.41 \\
Lingshu-7B ~\cite{xu2025lingshu}  & 12.00 & 30.58 & 25.77 & 27.48 & 20.50 & 30.94 & 48.00  & 20.00 & 27.08 \\
MedVLM-R1~\cite{pan2025medvlm} & 10.10 & 23.88 & 14.30 & 16.25 & 0.00 & 21.28 & 24.00  & 26.00 & 16.50 \\
Med-R1-Diagnosis ~\cite{lai2025medr1} & 6.30 & 22.93 & 8.91 & 20.13 & 2.00 & 19.01 & 16.00 &  22.00 & 15.29 \\
Chiron-o1-8B ~\cite{sunchiron} & 4.60 & 20.94 & 17.37 & 34.95 & 2.03 & 31.20 & 42.00 & 23.00 & 21.61 \\
MedGemma-27B ~\cite{sellergren2025medgemma} & 11.90 & 20.44 & 17.09 & 21.65 & 22.17 & 32.69 & 68.67 & 11.00 & 21.56 \\
HealthGPT~\cite{lin2025healthgpt} & 18.00 & 20.43 & 8.84 & 19.44 & 7.50 & 31.02 & 54.67 & 14.00 & 17.32 \\
OralGPT-Omni~\cite{hao2025oralgpt-omni} & \underline{66.80} & \textbf{56.60} & \underline{39.99} & \underline{48.11} & \underline{65.90} & \underline{56.01} & 47.33 & \underline{65.00} & \underline{51.84} \\
\hline
\rowcolor{mylightblue} \multicolumn{10}{l}{\textit{Medical Agents}} \\ 
\hline
MedRAX~\cite{fallahpour2025medrax} & 29.10 & 45.21 & 20.93 & 13.38 & 10.43 & 21.36 & 75.33 & 33.00 & 25.27 \\
MedAgents~\cite{tang2024medagents} & 2.50 & 41.61 & 22.59 & 39.11 & 0.00 & 38.33 & 0.00 & 19.00 & 29.52 \\
MMedAgent~\cite{li2024mmedagent} & 1.10 & 4.71 & 1.48 & 2.02 & 3.47 & 6.42 & 0.67 & 5.00 & 3.29 \\
MDAgents~\cite{kim2024mdagents} & 0.00 & 0.00 & 0.00 & 0.00 & 0.00 & 0.00 & 34.00 & 0.00 & 0.18 \\
\textbf{OralAgent (Ours)} & \textbf{71.20} & \underline{55.65} & \textbf{58.22} & \textbf{50.06} & \textbf{67.23} & \textbf{59.27} & \underline{78.67} & \textbf{65.00} & \textbf{57.70} \\
\bottomrule

\end{tabular}
% \begin{flushleft}
% \textsuperscript{1} https://openai.com/gpt-5/ \\
% \textsuperscript{2} https://openai.com/zh-Hant-HK/index/introducing-o3-and-o4-mini/ \\
% \textsuperscript{3} https://x.ai/grok/ \\
% \textsuperscript{4} https://www.anthropic.com/news/claude-sonnet-4-5/ \\ 
% \textsuperscript{5} https://www.volcengine.com/product/doubao/ \\
% \end{flushleft}
\end{table*}
\begin{table*}[t]
\centering
\caption{Performance on the MMOral-OPG benchmark for various MLLMs on open-ended VQA tasks. The best-performing model is highlighted \textbf{in-bold}, while the second-best is \underline{underlined}.
% \vspace{-0.2cm}
}
\label{tab:MMOral-OPG}
% \scriptsize
% \small
\renewcommand{\arraystretch}{1.0}
% \resizebox{0.48\textwidth}{!}{%
\resizebox{0.7\textwidth}{!}{%
\begin{tabular}{@{}l|ccccccc}
\toprule
\multirow{2}{*}{\textbf{Model}} 
  & \multicolumn{7}{c}{\textbf{Open‐ended VQA}} \\
\cline{2-8} 
\rule{0pt}{2.5ex} 
  & \textbf{Teeth} & \textbf{Patho} & \textbf{His} & \textbf{Jaw} & \textbf{Summ} & \textbf{Report} & \textbf{Overall} \\
\hline
\rowcolor{mylightblue} \multicolumn{8}{l}{\textit{General-purpose MLLMs}} \\ 
\hline
GPT-5~\cite{gpt5}  & \underline{39.77} & 29.32 & {44.05} & \textbf{78.56} & \underline{40.12} & 28.20 & {42.42} \\
GPT-4V~\cite{hurst2024gpt4v}  & {31.46} & 23.79 & 39.51 & 69.81 & {34.29} & \underline{43.70} & {39.38} \\
Gemini-2.5-Flash~\cite{team2023gemini} & 28.04 & 24.77 & 31.90 & 47.81 & 12.98 & 16.70 & 27.84 \\
Qwen-Max-VL~\cite{Qwen-VL} & 2.10 & 4.47 & 7.06 & 11.62 & 7.98 & 5.50 & 5.29 \\
Deepseek-VL-7b-chat~\cite{lu2024deepseek} & 16.48 & 7.50 & 13.44 & 34.56 & 9.52 & 9.60 & 15.95  \\
GLM-4V-9B-Thinking~\cite{glm2024chatglm}  & 20.94 & 9.70 & 18.77 & 26.62 & 12.74 & 21.30 & 19.74 \\
Qwen2.5-VL-72B~\cite{bai2025qwen25vl}  & 13.90 & 15.83 & 15.40 & 27.12 & 7.38 & 11.50 & 15.38  \\ \hline
\rowcolor{mylightblue} \multicolumn{8}{l}{\textit{Medical Specific MLLMs}} \\ \hline
LLaVA-Med~\cite{li2023llavamed}  & 0.91 & 1.52 & 0.00 & 0.00 & 0.00 & 24.50 & 4.76  \\
HealthGPT-XL32~\cite{lin2025healthgpt}  & 30.64 & 25.83 & 27.98 & 51.12 & 17.02 & 8.00 & 27.80 \\
MedVLM-R1~\cite{pan2025medvlm}  & 22.42 & 13.71 & 24.42 & 43.88 & 13.57 & 25.80 & 24.70 \\
MedDr~\cite{he2024meddr}  & 22.99 & {32.58} & 29.57 & 52.44 & 20.95 & 8.70 & 26.20  \\
OralGPT-Omni~\cite{hao2025oralgpt-omni} & {37.26} & \underline{43.94} & \underline{55.34} & {70.50} & {38.57} & 37.90 & \underline{45.31}  \\
\hline
\rowcolor{mylightblue} \multicolumn{8}{l}{\textit{Medical Agents}} \\ 
\hline
MedRAX~\cite{fallahpour2025medrax} & 32.35 & 18.11 & 42.64 & 59.63 & 45.36 & 32.10 & 36.73 \\
MedAgents~\cite{tang2024medagents} & 30.94 & 24.17 & 44.17 & 67.00 & 37.38 & 15.40 & 34.71 \\
MMedAgent & 16.00 & 10.08 & 15.15 & 39.81 & 8.93 & 0.50 & 15.19 \\
MDAgents~\cite{kim2024mdagents} & 36.19 & 31.52 & 54.36 & 64.94 & 39.29 & 26.90 & 40.50 \\
\textbf{OralAgent (Ours)} & \textbf{55.91} & \textbf{56.74} & \textbf{59.14} & \underline{76.25} & \textbf{70.00} & \textbf{61.00} & \textbf{61.00} \\
\bottomrule
\end{tabular}%
}
% \vspace{-0.3cm}
\end{table*}

\begin{table*}[t]
\centering
\renewcommand{\arraystretch}{1.25}
\setlength{\tabcolsep}{5pt}
\caption{Performance comparison of various individual LLM on the OralQA-ZH benchmark, as well as the improvements obtained when each LLM serves as the core orchestrator in OralAgent and performs RAG over the OralCorpus knowledge base. The use of OralAgent significantly improves performance. The best-performing model in each category is highlighted \textbf{in-bold}, while the second-best is \underline{underlined}.
}
\label{tab:rag_comparison}
\begin{tabular}{lcccccccccccc}
\toprule
\textbf{Model} & \textbf{Endo} & \textbf{Perio} & \textbf{OMFS} & \textbf{Prosth} & \textbf{Ortho} &
\textbf{OMD} & \textbf{PedDent} & \textbf{OMFR} & \textbf{PrevDent} & \textbf{OralEpi} & \textbf{OMFP} & \textbf{Overall} \\
\hline
\rowcolor{mylightblue} \multicolumn{13}{l}{\textbf{\textit{Individual LLM Baselines}}} \\
\hline
GPT-5.4~\cite{gpt5_4} & 74.44 & 89.89 & 76.11 & 66.07 & {37.50} & {90.74} & 82.14 & 58.33 & 63.41 & 75.86 & 82.50 & 74.69 \\
GPT-5-nano~\cite{gpt5} & 63.91 & 80.90 & 60.00 & 51.19 & 25.00 & 77.78 & 60.71 & 50.00 & 58.54 & 72.41 & 80.00 & 62.53 \\
Gemini-3.1-pro-thinking~\cite{team2023gemini}  & 55.64 & 71.91 & 69.44 & 31.55 & 0.00 & 90.74 & 71.43 & 58.33 & 19.51 & 3.45 & 12.50 & 50.88 \\
Kimi-k2.5~\cite{team2026kimi} & 62.41 & 65.17 & 59.44 & 57.74 & 29.17 & 75.93 & 82.14 & 75.00 & 68.29 & 34.48 & 70.00 & 61.53\\
MiniMax-M2.7-Guan~\cite{minimax27} & 57.89 & 62.92 & 56.11 & 50.00 & 25.00 & 59.26 & 53.57 & 33.33 & 34.15 & 48.28 & 57.50 & 53.38\\
GLM-4.7-Guan~\cite{v2507glm} & 41.35 & 22.47 & 18.89 & 17.86 & 0.00 & 9.26 & 14.29 & 33.33 & 34.15 & 55.17 & 40.00 & 24.81\\
Grok-4~\cite{grok4} & 73.68 & 85.39 & 77.78 & 65.48 & 45.83 & 81.48 & 78.57 & 58.33 & 60.98 & 82.76 & 80.00 & 73.81 \\
Qwen3.5-27B~\cite{qwen3.5} & 72.93 & 71.91 & 82.78 & 76.19 & 33.33 & 88.89 & 78.57 & 58.33 & 60.98 & 86.21 & 75.00 & 75.56 \\
HuatuoGPT-V-34B~\cite{zhang2023huatuogpt} & 58.65 & 67.42 & 47.78 & 45.83 & 25.00 & 61.11 & 35.71 & 33.33 & 51.22 & 62.07 & 72.50 & 52.88 \\
Lingshu-32B~\cite{xu2025lingshu} & 69.92 & 82.02 & 58.89 & 59.52 & 25.00 & 77.78 & 78.57 & 58.33 & 53.66 & 65.52 & 67.50 & 64.79 \\
MedDr-42B~\cite{he2024meddr} & 68.42 & 67.42 & 56.67 & 54.76 & 25.00 & 70.37 & 60.71 & 50.00 & 56.10 & 65.52 & 70.00 & 60.40\\
Hulu-Med-14B~\cite{jiang2025hulu} & 30.83 & 31.46 & 47.22 & 54.17 & 29.17 & 24.07 & 10.71 & 58.33 & 53.66 & 75.86 & 62.50 & 43.11 \\
\hline
\rowcolor{mylightblue} \multicolumn{13}{l}{\textbf{\textit{OralAgent with Different Core Orchestrators}}} \\
\hline
GPT-5.4~\cite{gpt5_4} & \textbf{81.20} & \textbf{92.13} & \textbf{85.00} & \textbf{77.98} & \textbf{41.67} & 88.89 & \underline{85.71} & {75.00} & \textbf{78.05} & {79.31} & \textbf{87.50} & \textbf{$82.08_{\color{mugreen}{+7.39}}$} \\
GPT-5-nano~\cite{gpt5} & 66.92 & 83.15 & 65.56 & 57.74 & 29.17 & 79.63 & 57.14 & 50.00 & {68.29} & 79.31 & \underline{82.50} & $66.92_{\color{mugreen}{+4.39}}$ \\
Gemini-3.1-pro-thinking~\cite{team2023gemini} & 74.53 & 88.05 & 81.71 & 49.52 & 34.0 & 86.05 & 82.07 & 41.04 & 44.39 & 57.01 & 53.08 & $69.02_{\color{mugreen}{+18.14}}$ \\ 
Kimi-k2.5~\cite{team2026kimi} & 82.71 & 85.39 & 74.44 & 63.69 & 33.33 & 87.04 & 82.14 & \textbf{83.33} & 60.98 & 62.07 & 67.50 & $73.31_{\color{mugreen}{+11.78}}$ \\ 
MiniMax-M2.7-Guan~\cite{minimax27} & \underline{81.20} & \underline{91.01} & \underline{84.44} & {69.05} & 25.00 & \textbf{92.59} & {85.71} & 75.00 & \underline{75.61} & \textbf{86.21} & 80.00 & $\underline{79.45}_{\color{mugreen}{+26.07}}$ \\ 
GLM-4.7-Guan~\cite{v2507glm} & 36.84 & 58.43 & 53.89 & 41.67 & 25.00 & 74.07 & 50.00 & 50.00 & 51.22 & 31.03 & 45.00 & $47.87_{\color{mugreen}{+23.06}}$ \\ 
Grok-4~\cite{grok4} & 78.20 & 86.52 & 81.11 & \underline{72.62} & 37.50 & \underline{90.74} & \textbf{89.29} & \underline{83.33} & 68.29 & 82.76 & 82.50 & $78.57_{\color{mugreen}{+4.76}}$ \\ 
Qwen3.5-27B~\cite{qwen3.5} & 79.70 & 92.13 & 80.00 & 71.43 & \underline{41.67} & 88.89 & 78.57 & 83.33 & 65.85 & \underline{86.21} & 77.50 & $78.32_{\color{mugreen}{+2.76}}$ \\ 
HuatuoGPT-V-34B~\cite{zhang2023huatuogpt} & 65.41 & 67.42 & 54.44 & 54.17 & 33.33 & 72.22 & 35.71 & 41.67 & 65.85 & 62.07 & 70.00 & $59.02_{\color{mugreen}{+6.14}}$ \\ 
Lingshu-32B~\cite{xu2025lingshu} & 74.44 & 87.64 & 73.89 & 61.90 & 25.00 & 83.33 & 71.43 & 83.33 & 70.73 & 65.52 & 67.50 & $71.43_{\color{mugreen}{+6.64}}$ \\ 
MedDr-42B~\cite{he2024meddr} & 71.43 & 77.53 & 64.44 & 55.95 & 25.00 & 81.48 & 57.14 & 75.00 & 63.41 & 75.86 & 70.00 & $65.79_{\color{mugreen}{+5.39}}$ \\ 
Hulu-Med-14B~\cite{jiang2025hulu} & 72.18 & 86.52 & 74.44 & 60.12 & 41.67 & 81.48 & 57.14 & 66.67 & 70.73 & 68.97 & 75.00 & $70.80_{\color{mugreen}{+27.69}}$ \\ 
\bottomrule
\end{tabular}
\end{table*}

\begin{table*}[!h]
\centering
\caption{ 
Ablation Study of the Input Instruction Comprehension Module in OralAgent: Prior Intent Recognition and Image Modality Classification. The best-performing model in each category is highlighted \textbf{in-bold}, while the second-best is \underline{underlined}. }
% \vspace{-5pt}
\label{tab:main}
\small
\renewcommand{\arraystretch}{1.1}
\setlength{\tabcolsep}{3.5pt}
\begin{tabular}{
    >{\raggedright\arraybackslash}p{3.0cm}  % Model
    >{\centering\arraybackslash}p{0.9cm}    % Intent
    >{\centering\arraybackslash}p{1.1cm}    % Modality
    *{9}{>{\centering\arraybackslash}p{1.15cm}} % 后面九列
}
\toprule
\multirow{2}{*}{Model} 
& \multirow{2}{*}{Intent} 
& \multirow{2}{*}{Modality} 
& \multicolumn{3}{c}{II} 
& \multirow{2}{*}{PA} 
& \multirow{2}{*}{CE} 
& \multirow{2}{*}{PI} 
& \multirow{2}{*}{TP}
& \multirow{2}{*}{IV}
& \multirow{2}{*}{Overall} \\
\cmidrule(lr){4-6}
& & & Loc & Dx-I & Dx-R &  &  &  &  &  &  \\
\hline

\textbf{OralAgent (Ours)}      & $\checkmark$ & $\checkmark$
& \textbf{71.20} & \textbf{55.65} & \underline{58.22} & \textbf{50.06} & \textbf{67.23} & \underline{59.27} & \textbf{78.67} & \textbf{65.00} & \textbf{57.70} \\

w/o Intent              & $\times$     & $\checkmark$
& \underline{68.50} & \underline{53.01} & \textbf{59.10} & \underline{48.78} & 65.20 & \textbf{59.71} & \underline{74.67} & 45.00 & \underline{56.66} \\

w/o Modality            & $\checkmark$ & $\times$
& 61.80 & 51.78 & 48.42 & 47.74 & 64.67 & 57.47 & 73.33 & \underline{56.00} & 52.76 \\

w/o Intent \& Modality  & $\times$     & $\times$
& 59.90 & 49.47 & 46.46 & 47.35 & \underline{65.63} & 57.18 & 67.33 & 47.00 & 51.50 \\

\bottomrule
\end{tabular}
\end{table*}

\begin{figure*}[!ht]
\centering
\includegraphics[width=1.0\linewidth]{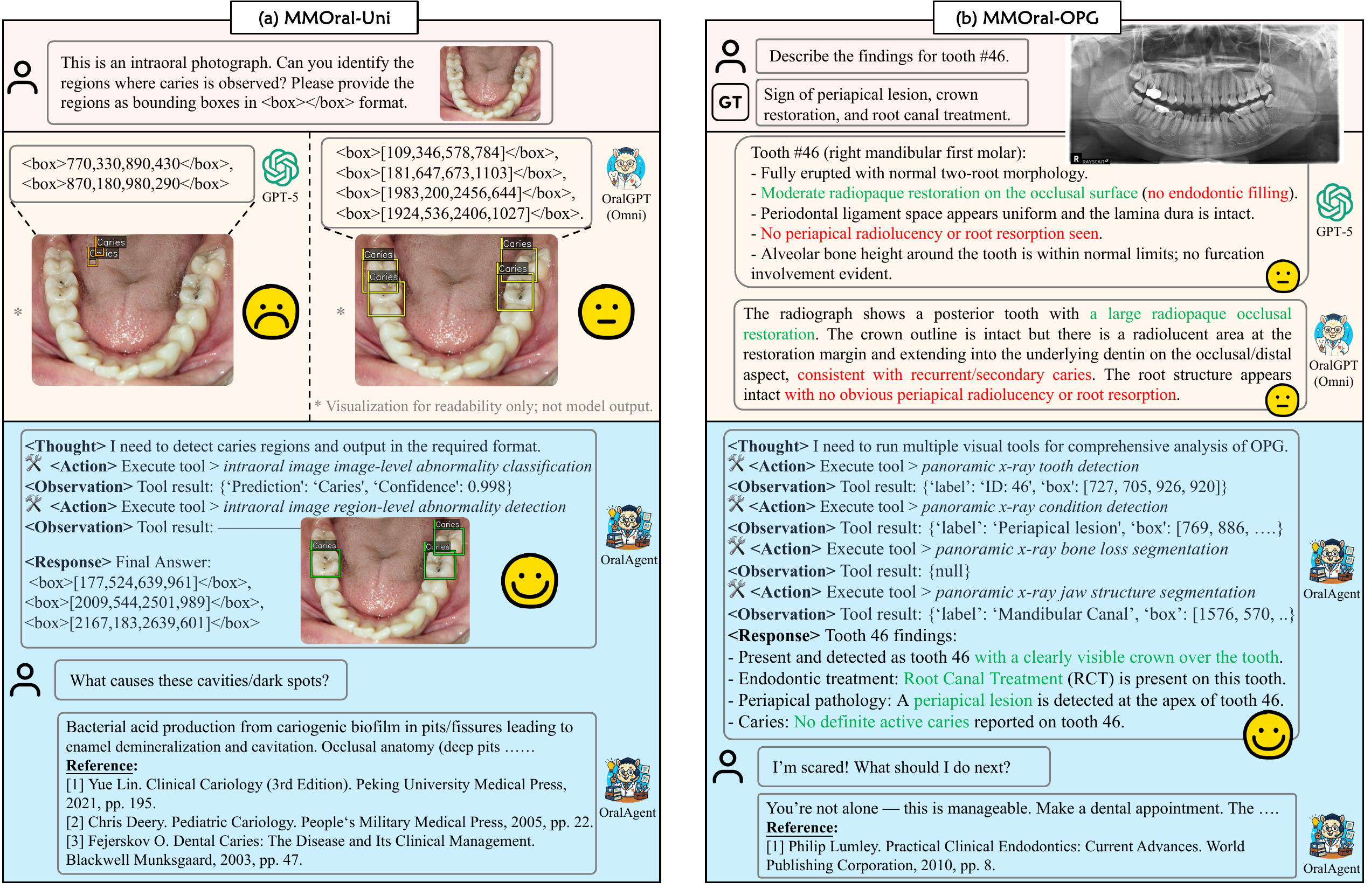}
\caption{Case study on MMOral-Uni and MMOral-OPG benchmarks, respectively. (a) A case about an intraoral image from the MMOral-Uni benchmark. 
% GPT-5 fails to locate the carious region, and OralGPT-Omni identifies an approximate area but generates imprecise bounding boxes. 
Note that the visualization is created through manual post-processing for readability only. Our OralAgent accurately localizes the carious region by invoking tools, and it can directly generate visualized outputs that support intuitive user interpretation. (b) 
A case about a panoramic x-ray from the MMOral-OPG benchmark. OralAgent performs a comprehensive assessment using four dental-specialized tools and successfully identifies the full condition of tooth 46. Besides, OralAgent provides traceable references throughout multi-turn interactions, further ensuring its reliability.
}
\label{fig:case_study}
\end{figure*}

\begin{figure*}[!ht]
\centering
\includegraphics[width=1.0\linewidth]{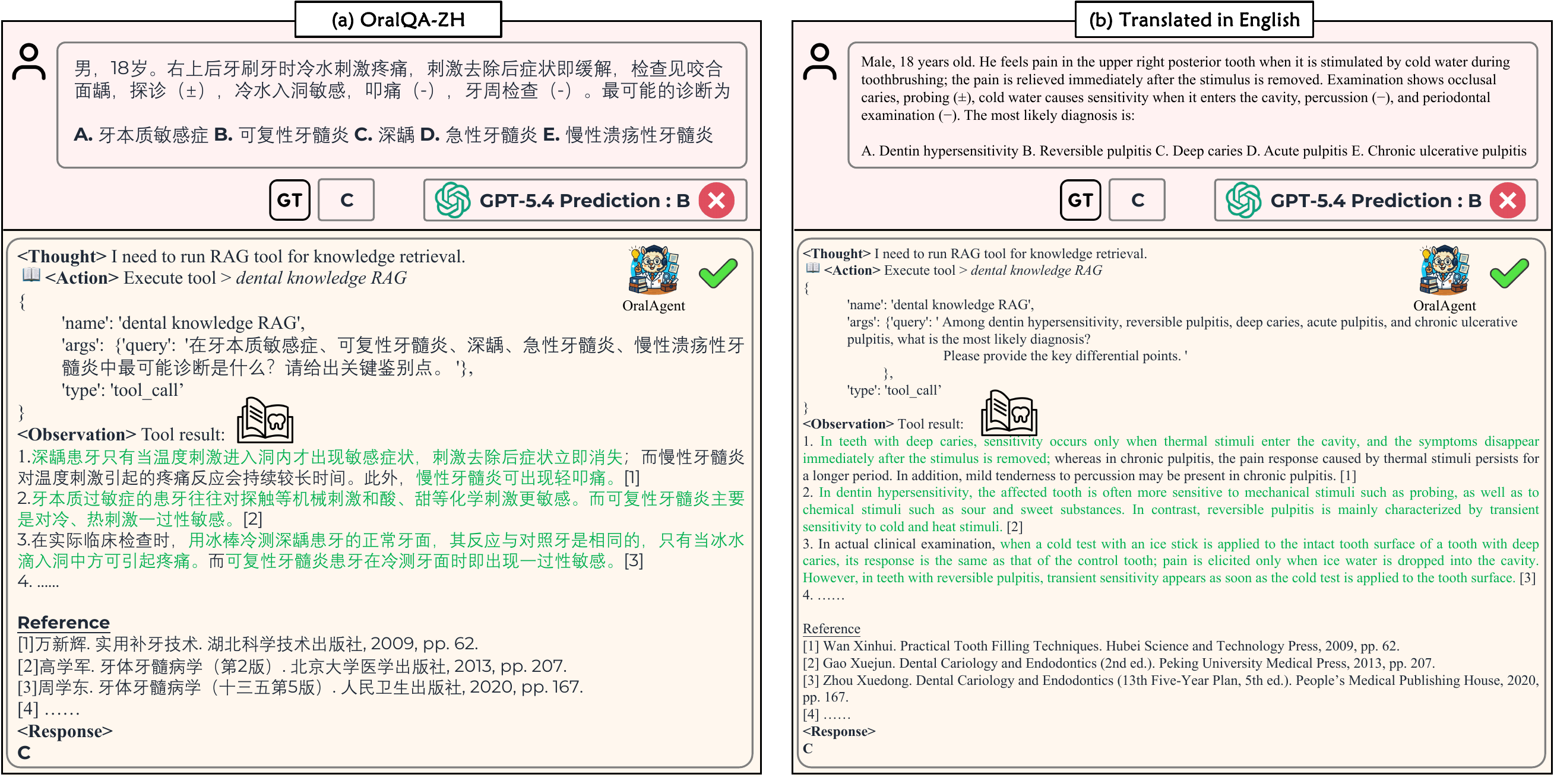}
\caption{
Case study on the OralQA-ZH benchmark. (a) Original example in Chinese. OralAgent can automatically invoke RAG tools to retrieve knowledge relevant to the question, generate the correct answer, and provide precise supporting references. (b) English translation for readability.
}
\label{fig:case_study_OralQA}
\end{figure*}

\subsection{Experimental Setup}
\noindent
\textbf{Benchmarks.}
We evaluated our OralAgent on two public dental multimodal benchmarks and one proposed text-only benchmark.
\begin{itemize}
    \item \textbf{MMOral-Uni Benchmark.} It is a unified multimodal benchmark for dental image analysis, and it comprises 2,809 open-ended QA pairs to simulate realistic user interactions. The dental imaging modalities cover intraoral photographs, periapical and cephalometric radiographs, pathological images, intraoral videos, and interleaved image-text inputs, offering a comprehensive evaluation suite for MLLMs in digital dentistry.
    \item \textbf{MMOral-OPG Benchmark.} It is a multimodal benchmark for panoramic X-ray analysis, and it has 578 open-ended questions. This benchmark covers five key diagnostic dimensions, including the condition of teeth, pathological findings, historical treatments, jawbone observations, and clinical summary \& recommendations. 
    \item \textbf{OralQA Benchmark.} Our proposed simplified Chinese text-only benchmark containing 798 multiple-choice questions across 11 oral subspecialties. Model performance is measured by accuracy across all questions. 
    
\end{itemize}

\noindent
\textbf{Baselines.}
Regarding multimodal benchmarks, we evaluated OralAgent against four types of models, including proprietary MLLMs, open-source MLLMs, medical MLLMs, and medical agents. We evaluate 6 proprietary MLLMs via API: GPT-5~\cite{gpt5}, o3~\cite{gpto3}, Grok-4~\cite{grok4}, Doubao-1.5-vision~\cite{doubao}, Gemini-2.5-flash~\cite{team2023gemini}, and GLM-4.5v~\cite{glm2024chatglm}. For Open-Source MLLMs, we test 12 powerful models, including Qwen2.5-VL-7B~\cite{Qwen-VL}, Qwen3-VL-8B-Instruct~\cite{Qwen-VL}, Qwen3-VL-235B-A22B~\cite{Qwen-VL}, GLM-4.1V-9B-Thinking~\cite{glm2024chatglm}, InternVL3.5-8B~\cite{wang2025internvl3}, LLaVA-v1.6-Mistral-7B~\cite{llava16}, LLaVA-OneVision~\cite{li2024llavaonevision}, MiMo-VL-7B~\cite{xiaomi2025mimo}, Phi-4-multimodal-instruct~\cite{abdin2024phi}, Mistral-Small-3.1-24B~\cite{Mistral}, R-4B~\cite{yang2025r4b}, and Ovis2.5-9B~\cite{lu2025ovis2}. In addition, we evaluate 8 representative medical MLLMs: LLaVA-Med~\cite{li2023llavamed}, HuatuoGPT-Vision-7~\cite{chen2024huatuogpt}, Lingshu-7B~\cite{xu2025lingshu}, MedVLM-R1~\cite{pan2025medvlm}, Med-R1-Diagnosis~\cite{lai2025medr1}, Chiron-o1-8B~\cite{sunchiron}, MedGemma-27B~\cite{sellergren2025medgemma}, HealthGP~\cite{lin2025healthgpt}, and OralGPT-Omni~\cite{hao2025oralgpt-omni}. We also evaluate 4 medical agent systems, including MDAgents~\cite{kim2024mdagents}, MedAgents~\cite{tang2024medagents}, MMedAgent~\cite{li2024mmedagent}, and MedRAX~\cite{fallahpour2025medrax}. For the text-only OralQA-ZH benchmark, we evaluated twelve LLMs, including GPT-5.4~\cite{gpt5_4}, GPT-5-nano~\cite{gpt5}, Gemini-3.1-pro-thinking~\cite{team2023gemini}, Kimi-k2.5~\cite{team2026kimi}, MiniMax-M2.7-Guan~\cite{minimax27}, GLM-4.7-Guan~\cite{v2507glm}, Grok-4, Qwen3.5-27B, HuatuoGPT-V-34B~\cite{zhang2023huatuogpt}, Lingshu-32B~\cite{xu2025lingshu}, MedDr-42B~\cite{he2024meddr}, and Hulu-Med 14B~\cite{jiang2025hulu}. We evaluated both the models' paramatric performance and the improvements brought by OralAgent's RAG module based on the OralCorpus knowledge base. Each LLM can function as the core orchestrator of OralAgent and invoke RAG-based tools for external knowledge retrieval in order to answer knowledge-intensive questions in OralQA-ZH benchmark. The default Top-k in the OralAgent's RAG module was set to 7.

\subsection{Quantitative Analysis}
\noindent
\textbf{MMOral-Uni Benchmark.}
As shown in Table~\ref{tab:mmoral-uni}, OralAgent achieves a new state-of-the-art overall score of 57.70, surpassing the dental-specialized MLLM OralGPT Omni by 5.86 points. OralAgent consistently outperforms a wide range of proprietary MLLMs, open-source MLLMs, medical MLLMs, and medical agents, demonstrating strong capability in dental image analysis across six modalities. Notably, through autonomous tool invocation, OralAgent exhibits substantial improvements on localization-sensitive tasks, including Loc, Dx-R, CE, and PI, which benefit from the precise positional information provided by dental visual tools. Figure~\ref{fig:tool_usage} illustrates the distribution of tool invocations made by OralAgent on the MMOral-Uni benchmark, where the OralGPT-Omni model is invoked most frequently. On average, 3.67 tool calls are made per case, indicating OralAgent’s flexible tool selection and comprehensive tool-driven analysis.

\noindent
\textbf{MMOral-OPG Benchmark.}
The MMOral-OPG Benchmark results are presented in Table~\ref{tab:MMOral-OPG}, which compares the performance of OralAgent with general-purpose MLLMs, medical-specific MLLMs, and medical agents. OralAgent surpasses the dental-specialized MLLM OralGPT-Omni by 15.69 points and achieves a new state-of-the-art overall score of 61.00. At the same time, OralAgent exceeds the best-performing medical agent, MDAgents, by 20.5 points, demonstrating its clear superiority in panoramic x-ray analysis.

\noindent
\textbf{OralQA-ZH Benchmark.}
We evaluate a range of existing LLMs on the OralQA-ZH benchmark, both with and without the dental RAG module in OralAgent. The results are presented in Table~\ref{tab:rag_comparison}. Overall, integrating OralAgent's RAG module substantially improves model accuracy on multidisciplinary QA tasks in the dental domain. Among all individual LLM baselines, GPT-5.4 achieved the highest score of 74.69. When GPT-5.4 was used as the core orchestrator with access to the RAG tool, its score increased to 82.08, yielding an improvement of 7.39 points. This result confirms the effectiveness of both the RAG module in OralAgent and the proposed OralCorpus knowledge base. MiniMax-M2.7-Guan~\cite{minimax27} and Hulu-Med 14B~\cite{jiang2025hulu} also exhibit marked performance improvements after incorporating the RAG module, with respective gains of 26.07 and 27.69 points. This finding suggests that the RAG module and the OralCorpus knowledge base can effectively provide specialized dental knowledge that the model does not inherently possess, thereby enhancing its overall performance in the dental domain.

\subsection{Case Studies}
\noindent
\textbf{Dental Image Analysis.}
We present two representative cases from the MMOral‑Uni and MMOral‑OPG benchmarks, shown in Figure~\ref{fig:case_study}, to compare the performance of OralAgent with GPT‑5 and OralGPT‑Omni. In Figure~\ref{fig:case_study}(a), the task involves identifying dental caries in an intraoral image, a scenario that reflects common user needs in daily scenarios. GPT‑5 fails to locate the carious region and only follows basic formatting instructions. OralGPT‑Omni identifies an approximate area but generates imprecise bounding boxes. It is important to note that both GPT‑5 and OralGPT‑Omni output only numerical coordinates; the visualization in Figure~\ref{fig:case_study}(a) is created through manual post‑processing and is not produced directly by the models. In contrast, OralAgent accurately localizes the carious region by invoking tools for image‑level abnormality classification and region‑level detection, and it can directly generate visualized outputs that support intuitive user interpretation. During subsequent multi‑turn interactions, OralAgent also retrieves information from 368 dental textbooks through a RAG‑based workflow and produces responses with traceable citations, including book titles and page numbers, thereby improving the reliability and credibility of its outputs.
Figure~\ref{fig:case_study}(b) illustrates a case concerning the condition of tooth 46 in a panoramic x‑ray image, which aligns with routine clinical diagnostic needs. Neither GPT‑5 nor OralGPT‑Omni detects the periapical lesion. OralAgent, however, performs a comprehensive assessment of the panoramic x‑ray by orchestrating multiple dental-specialized tools and successfully identifies the full condition of tooth 46. Specifically, OralAgent first invokes a tooth‑detection tool to localize tooth 46, and then sequentially calls three specialized models to evaluate (1) pathological abnormalities of tooth 46, (2) the extent of surrounding bone loss, and (3) the status of the mandibular canal, thereby achieving a thorough condition assessment. As in the previous case, OralAgent provides traceable references throughout the multi‑turn interaction, further enhancing the reliability and trustworthiness of its responses.

\noindent
\textbf{Dental Knowledge-based Diagnosis.} 
We further analyze the complete problem-solving process of OralAgent on a knowledge-intensive, text-only query, using a case from OralQA-ZH, as illustrated in Figure~\ref{fig:case_study_OralQA}. The figure presents both the original Chinese dialogue and its English translation for readability. For the GPT-5.4 model, if it relies solely on its own parametric knowledge, it produces the incorrect answer, option B. In contrast, when OralAgent adopts GPT-5.4 as the core orchestrator, it first invokes the RAG tool and automatically generates the JSON-formatted parameters and query required for retrieval based on the user’s question. The RAG tool then returns multiple knowledge snippets and references most relevant to the question. Finally, OralAgent answers the question using the retrieved knowledge as supporting evidence, yielding the correct answer, option C. Benefiting from RAG and the OralCorpus knowledge base, OralAgent is not only capable of answering users’ knowledge-related questions, but also of providing precise reasoning traces and reference sources, thereby enhancing the reliability and explainability of the conversational intelligence system.

\subsection{Ablation Studies}
The success of OralAgent depends on the core orchestrator’s ability to perform reasoning and planning based on user instructions. Its understanding of user intent, together with its interpretation of uploaded image modalities, largely determines the quality of reasoning and the correctness of tool selection and invocation. To assess these factors, we conducted ablation studies on the input instruction comprehension module to examine how prior intent recognition and image modality classification influence performance in dental image analysis. The results are summarized in Table 5. Using the full OralAgent system as the baseline, removing only the intent recognition module reduced the overall score on the MMOral-Uni benchmark from 57.70 to 56.66. Removing only the modality classification module produced a larger drop, decreasing the score from 57.70 to 52.76. This finding indicates that explicit image modality information is crucial for effective tool selection and invocation by the orchestrator. Finally, removing both the intent recognition and modality classification modules resulted in a 6.2-point decrease in overall performance, underscoring the essential role of the input instruction comprehension module in OralAgent.

\section{Conclusion}
OralAgent establishes a new paradigm for dental AI by dynamically and autonomously performing reasoning, planning, tool invocation, and knowledge retrieval, enabling more intelligent and reliable interactions within digital workflows. By integrating 22 tools and 368 dental textbooks, OralAgent provides a robust foundation for dental agent applications. Extensive experiments demonstrate its superior performance in dental image analysis across multiple modalities and in knowledge comprehension across diverse oral subspecialties, underscoring the advantages of explicit step‑by‑step reasoning in dental AI. We believe that OralAgent will help pave the way for the next generation of dental AI and accelerate its practical clinical adoption. Future work will focus on expanding the toolkit to include understanding of 3D imaging modalities and upgrading the RAG module into a fully multimodal RAG branch. We are confident that these advancements will further enhance the clinical value and real‑world impact of OralAgent.

\appendices

% \section*{Appendix and the Use of Supplemental Files}
% Appendices, if needed, appear before the acknowledgment. If an appendix is not
% critical to the main message of the manuscript and is included only for thoroughness
% or for reader reference, then consider submitting appendices as supplemental materials.
% Supplementary files are available to readers through IEEE \emph{Xplore\textregistered}
% at no additional cost to the authors but they do not appear in print versions.
% Supplementary files must be uploaded in ScholarOne as supporting documents, but for
% accepted papers they should be uploaded as Multimedia documents. Refer readers
% to the supplementary files where appropriate within the manuscript text using footnotes.
% \footnote{Supplementary materials are available in the supporting documents/multimedia tab.
% Further instructions on footnote usage are in the Footnotes section on the next page.}

% \section*{Acknowledgment}
% The preferred spelling of the word ``acknowledgment'' in American English is 
% without an ``e'' after the ``g.'' Use the singular heading even if you have 
% many acknowledgments. Avoid expressions such as ``One of us (S.B.A.) would 
% like to thank $\ldots$ .'' Instead, write ``F. A. Author thanks $\ldots$ .'' In most 
% cases, sponsor and financial support acknowledgments are placed in the 
% unnumbered footnote on the first page, not here.

\bibliographystyle{IEEEtran}
\bibliography{OralAgent}

\begin{thebibliography}{10}
\providecommand{\url}[1]{#1}
\csname url@samestyle\endcsname
\providecommand{\newblock}{\relax}
\providecommand{\bibinfo}[2]{#2}
\providecommand{\BIBentrySTDinterwordspacing}{\spaceskip=0pt\relax}
\providecommand{\BIBentryALTinterwordstretchfactor}{4}
\providecommand{\BIBentryALTinterwordspacing}{\spaceskip=\fontdimen2\font plus
\BIBentryALTinterwordstretchfactor\fontdimen3\font minus \fontdimen4\font\relax}
\providecommand{\BIBforeignlanguage}[2]{{%
\expandafter\ifx\csname l@#1\endcsname\relax
\typeout{** WARNING: IEEEtran.bst: No hyphenation pattern has been}%
\typeout{** loaded for the language `#1'. Using the pattern for}%
\typeout{** the default language instead.}%
\else
\language=\csname l@#1\endcsname
\fi
#2}}
\providecommand{\BIBdecl}{\relax}
\BIBdecl

\bibitem{hao2024semi}
J.~Hao, L.~M. Wong, Z.~Shan, Q.~Y.~H. Ai, X.~Shi, J.~K.~H. Tsoi, and K.~F. Hung, ``A semi-supervised transformer-based deep learning framework for automated tooth segmentation and identification on panoramic radiographs,'' \emph{Diagnostics}, vol.~14, no.~17, p. 1948, 2024.

\bibitem{hao2024semit}
J.~Hao, M.~Liu, L.~He, L.~Yao, J.~K.~H. Tsoi, and K.~F. Hung, ``Semit-sam: Building a visual foundation model for tooth instance segmentation on panoramic radiographs,'' in \emph{International Conference on Medical Image Computing and Computer-Assisted Intervention}.\hskip 1em plus 0.5em minus 0.4em\relax Springer, 2024, pp. 110--121.

\bibitem{hao2024tmamba}
J.~Hao, Y.~Zhu, L.~He, M.~Liu, J.~K.~H. Tsoi, and K.~F. Hung, ``T-mamba: a unified framework with long-range dependency in dual-domain for 2d \& 3d tooth segmentation,'' \emph{arXiv preprint arXiv:2404.01065}, 2024.

\bibitem{hao2025oraldataset}
J.~Hao, A.~Nalley, A.~W.~K. Yeung, R.~Tanaka, Q.~Y.~H. Ai, W.~Y.~H. Lam, Z.~Shan, Y.~Y. Leung, A.~AlHadidi, M.~M. Bornstein \emph{et~al.}, ``Characteristics, licensing, and ethical considerations of openly accessible oral-maxillofacial imaging datasets: a systematic review,'' \emph{npj Digital Medicine}, vol.~8, no.~1, p. 412, 2025.

\bibitem{hao2026photography}
J.~Hao, K.~Guo, Z.~Shan, Y.~Yang, J.~K.-H. Tsoi, P.~P.~Y. Lam, and K.~F. Hung, ``Photography-based dental plaque detection and report generation among preschool children using plaquesam,'' 2026.

\bibitem{wu2024cephalometric}
H.~Wu, C.~Wang, L.~Mei, T.~Yang, M.~Zhu, D.~Shen, and Z.~Cui, ``Cephalometric landmark detection across ages with prototypical network,'' in \emph{International conference on medical image computing and computer-assisted intervention}.\hskip 1em plus 0.5em minus 0.4em\relax Springer, 2024, pp. 155--165.

\bibitem{guan2026high}
J.~Guan, J.~Guo, Q.~Chen, J.~Chen, Y.~Cai, Y.~He, Z.~Huang, Y.~Wang, and Y.~Xie, ``A high magnifications histopathology image dataset for oral squamous cell carcinoma diagnosis and prognosis,'' \emph{Scientific Data}, 2026.

\bibitem{sun2024descriptive}
Y.~Sun, J.~Hao, K.~Zhu, J.-J. Liu, Y.~Zhao, X.~Li, G.~Zhang, Z.~Li, and J.~Wang, ``Descriptive caption enhancement with visual specialists for multimodal perception,'' \emph{arXiv preprint arXiv:2412.14233}, 2024.

\bibitem{hao2024fullanno}
J.~Hao, Y.~Zhao, S.~Chen, Y.~Sun, Q.~Chen, G.~Zhang, K.~Yao, E.~Ding, and J.~Wang, ``Fullanno: A data engine for enhancing image comprehension of mllms,'' \emph{arXiv preprint arXiv:2409.13540}, 2024.

\bibitem{hao2025oralgpt-omni}
J.~Hao, Y.~Liang, L.~Lin, Y.~Fan, W.~Zhou, K.~Guo, Z.~Ye, Y.~Sun, X.~Zhang, Y.~Yang \emph{et~al.}, ``Oralgpt-omni: A versatile dental multimodal large language model,'' \emph{CVPR}, 2026.

\bibitem{cai2025dentalgpt}
Z.~Cai, J.~Zhang, J.~Zhao, Z.~Zeng, Y.~Li, J.~Liang, J.~Chen, Y.~Yang, J.~You, S.~Deng \emph{et~al.}, ``Dentalgpt: Incentivizing multimodal complex reasoning in dentistry,'' \emph{arXiv preprint arXiv:2512.11558}, 2025.

\bibitem{liu2025performance1}
Z.~Liu, A.~Nalley, J.~Hao, Q.~Y. H~Ai, A.~W. Kan~Yeung, R.~Tanaka, and K.~F. Hung, ``The performance of large language models in dentomaxillofacial radiology: a systematic review,'' \emph{Dentomaxillofacial Radiology}, p. twaf060, 2025.

\bibitem{yao2022react}
S.~Yao, J.~Zhao, D.~Yu, N.~Du, I.~Shafran, K.~R. Narasimhan, and Y.~Cao, ``React: Synergizing reasoning and acting in language models,'' in \emph{The eleventh international conference on learning representations}, 2022.

\bibitem{hao2025mmoral}
J.~Hao, Y.~Fan, Y.~Sun, K.~Guo, L.~Lin, J.~Yang, Q.~Y.~H. Ai, L.~M. Wong, H.~Tang, and K.~F. Hung, ``Towards better dental ai: A multimodal benchmark and instruction dataset for panoramic x-ray analysis,'' \emph{NeurIPS 2025}, 2025.

\bibitem{arslan2024rag}
M.~Arslan, H.~Ghanem, S.~Munawar, and C.~Cruz, ``A survey on rag with llms,'' \emph{Procedia computer science}, vol. 246, pp. 3781--3790, 2024.

\bibitem{kim2024mdagents}
Y.~Kim, C.~Park, H.~Jeong, Y.~S. Chan, X.~Xu, D.~McDuff, H.~Lee, M.~Ghassemi, C.~Breazeal, and H.~W. Park, ``Mdagents: An adaptive collaboration of llms for medical decision-making,'' \emph{Advances in Neural Information Processing Systems}, vol.~37, pp. 79\,410--79\,452, 2024.

\bibitem{li2024mmedagent}
B.~Li, T.~Yan, Y.~Pan, J.~Luo, R.~Ji, J.~Ding, Z.~Xu, S.~Liu, H.~Dong, Z.~Lin \emph{et~al.}, ``Mmedagent: Learning to use medical tools with multi-modal agent,'' in \emph{Findings of the Association for Computational Linguistics: EMNLP 2024}, 2024, pp. 8745--8760.

\bibitem{wang2025medagentpro}
Z.~Wang, J.~Wu, L.~Cai, C.~H. Low, X.~Yang, Q.~Li, and Y.~Jin, ``Medagent-pro: Towards evidence-based multi-modal medical diagnosis via reasoning agentic workflow,'' \emph{arXiv preprint arXiv:2503.18968}, 2025.

\bibitem{zhao2025medrag}
X.~Zhao, S.~Liu, S.-Y. Yang, and C.~Miao, ``Medrag: Enhancing retrieval-augmented generation with knowledge graph-elicited reasoning for healthcare copilot,'' in \emph{Proceedings of the ACM on Web Conference 2025}, 2025, pp. 4442--4457.

\bibitem{fallahpour2025medrax}
A.~Fallahpour, J.~Ma, A.~Munim, H.~Lyu, and B.~Wang, ``Medrax: Medical reasoning agent for chest x-ray,'' \emph{arXiv preprint arXiv:2502.02673}, 2025.

\bibitem{fan2026oralgpt-plus}
Y.~Fan, J.~Hao, H.~Chen, J.~Bao, Y.~Shao, Y.~Liang, K.~F. Hung, and H.~Tang, ``Oralgpt-plus: Learning to use visual tools via reinforcement learning for panoramic x-ray analysis,'' \emph{CVPR 2026}, 2026.

\bibitem{meng2025dentvlm}
Z.~Meng, J.~Hao, X.~Dai, Y.~Feng, J.~Liu, B.~Feng, H.~Wu, X.~Gai, H.~Zhu, T.~Hu \emph{et~al.}, ``Dentvlm: A multimodal vision-language model for comprehensive dental diagnosis and enhanced clinical practice,'' \emph{arXiv preprint arXiv:2509.23344}, 2025.

\bibitem{yu2026opgagent}
Z.~Yu, L.~Yang, B.~Babicka, M.~Hu, J.~Hao, A.~Huang, J.~Huang, Y.~Jin, J.~Wu, and Z.~Ge, ``Opgagent: An agent for auditable dental panoramic x-ray interpretation,'' \emph{arXiv preprint arXiv:2603.00462}, 2026.

\bibitem{wang2025prompt}
J.~Wang, E.~Shi, S.~Yu, Z.~Wu, H.~Hu, C.~Ma, H.~Dai, Q.~Yang, Y.~Kang, J.~Wu \emph{et~al.}, ``Prompt engineering for healthcare: Methodologies and applications,'' \emph{Meta-Radiology}, p. 100190, 2025.

\bibitem{rodriguez2024intentgpt}
J.~A. Rodriguez, N.~Botzer, D.~Vazquez, C.~Pal, M.~Pedersoli, and I.~Laradji, ``Intentgpt: Few-shot intent discovery with large language models,'' \emph{arXiv preprint arXiv:2411.10670}, 2024.

\bibitem{yang2025qwen3}
A.~Yang, A.~Li, B.~Yang, B.~Zhang, B.~Hui, B.~Zheng, B.~Yu, C.~Gao, C.~Huang, C.~Lv \emph{et~al.}, ``Qwen3 technical report,'' \emph{arXiv preprint arXiv:2505.09388}, 2025.

\bibitem{zhang2023biomedclip}
S.~Zhang, Y.~Xu, N.~Usuyama, H.~Xu, J.~Bagga, R.~Tinn, S.~Preston, R.~Rao, M.~Wei, N.~Valluri \emph{et~al.}, ``Biomedclip: a multimodal biomedical foundation model pretrained from fifteen million scientific image-text pairs,'' \emph{arXiv preprint arXiv:2303.00915}, 2023.

\bibitem{simeoni2025dinov3}
O.~Sim{\'e}oni, H.~V. Vo, M.~Seitzer, F.~Baldassarre, M.~Oquab, C.~Jose, V.~Khalidov, M.~Szafraniec, S.~Yi, M.~Ramamonjisoa \emph{et~al.}, ``Dinov3,'' \emph{arXiv preprint arXiv:2508.10104}, 2025.

\bibitem{zhang2022dino}
H.~Zhang, F.~Li, S.~Liu, L.~Zhang, H.~Su, J.~Zhu, L.~M. Ni, and H.-Y. Shum, ``Dino: Detr with improved denoising anchor boxes for end-to-end object detection,'' \emph{arXiv preprint arXiv:2203.03605}, 2022.

\bibitem{li2023maskdino}
F.~Li, H.~Zhang, H.~Xu, S.~Liu, L.~Zhang, L.~M. Ni, and H.-Y. Shum, ``Mask dino: Towards a unified transformer-based framework for object detection and segmentation,'' in \emph{Proceedings of the IEEE/CVF conference on computer vision and pattern recognition}, 2023, pp. 3041--3050.

\bibitem{wang2024mineru}
B.~Wang, C.~Xu, X.~Zhao, L.~Ouyang, F.~Wu, Z.~Zhao, R.~Xu, K.~Liu, Y.~Qu, F.~Shang \emph{et~al.}, ``Mineru: An open-source solution for precise document content extraction,'' \emph{arXiv preprint arXiv:2409.18839}, 2024.

\bibitem{zhang2025qwen3_embed_rerank}
Y.~Zhang, M.~Li, D.~Long, X.~Zhang, H.~Lin, B.~Yang, P.~Xie, A.~Yang, D.~Liu, J.~Lin \emph{et~al.}, ``Qwen3 embedding: Advancing text embedding and reranking through foundation models,'' \emph{arXiv preprint arXiv:2506.05176}, 2025.

\bibitem{gpt5}
\BIBentryALTinterwordspacing
OpenAI, ``Gpt-5,'' 2025. [Online]. Available: \url{https://openai.com/zh-Hans-CN/index/introducing-gpt-5}
\BIBentrySTDinterwordspacing

\bibitem{gpto3}
\BIBentryALTinterwordspacing
------, ``o3,'' 2025. [Online]. Available: \url{https://openai.com/zh-Hans-CN/index/introducing-o3-and-o4-mini}
\BIBentrySTDinterwordspacing

\bibitem{grok4}
\BIBentryALTinterwordspacing
xAI, ``grok4,'' 2025. [Online]. Available: \url{https://x.ai/grok}
\BIBentrySTDinterwordspacing

\bibitem{doubao}
\BIBentryALTinterwordspacing
ByteDance, ``Doubao-1.5-vision-pro,'' 2025. [Online]. Available: \url{https://www.volcengine.com/product/doubao}
\BIBentrySTDinterwordspacing

\bibitem{team2023gemini}
G.~Team, R.~Anil, S.~Borgeaud, J.-B. Alayrac, J.~Yu, R.~Soricut, J.~Schalkwyk, A.~M. Dai, A.~Hauth, K.~Millican \emph{et~al.}, ``Gemini: a family of highly capable multimodal models,'' \emph{arXiv preprint arXiv:2312.11805}, 2023.

\bibitem{v2507glm}
V.~Team, W.~Hong, W.~Yu, X.~Gu, G.~Wang, G.~Gan, H.~Tang, J.~Cheng, J.~Qi, J.~Ji \emph{et~al.}, ``Glm-4.5 v and glm-4.1 v-thinking: Towards versatile multimodal reasoning with scalable reinforcement learning, 2025,'' \emph{https://arxiv.org/abs/2507.01006}, 2025.

\bibitem{bai2025qwen25vl}
S.~Bai, K.~Chen, X.~Liu, J.~Wang, W.~Ge, S.~Song, K.~Dang, P.~Wang, S.~Wang, J.~Tang \emph{et~al.}, ``Qwen2. 5-vl technical report,'' \emph{arXiv preprint arXiv:2502.13923}, 2025.

\bibitem{wang2025internvl3}
W.~Wang, Z.~Gao, L.~Gu, H.~Pu, L.~Cui, X.~Wei, Z.~Liu, L.~Jing, S.~Ye, J.~Shao \emph{et~al.}, ``Internvl3. 5: Advancing open-source multimodal models in versatility, reasoning, and efficiency,'' \emph{arXiv preprint arXiv:2508.18265}, 2025.

\bibitem{llava16}
\BIBentryALTinterwordspacing
H.~Liu, C.~Li, Y.~Li, B.~Li, Y.~Zhang, S.~Shen, and Y.~J. Lee, ``Llava-next: Improved reasoning, ocr, and world knowledge,'' January 2024. [Online]. Available: \url{https://llava-vl.github.io/blog/2024-01-30-llava-next/}
\BIBentrySTDinterwordspacing

\bibitem{li2024llavaonevision}
B.~Li, Y.~Zhang, D.~Guo, R.~Zhang, F.~Li, H.~Zhang, K.~Zhang, P.~Zhang, Y.~Li, Z.~Liu \emph{et~al.}, ``Llava-onevision: Easy visual task transfer,'' \emph{arXiv preprint arXiv:2408.03326}, 2024.

\bibitem{xiaomi2025mimo}
L.~Xiaomi, B.~Xia, B.~Shen, D.~Zhu, D.~Zhang, G.~Wang, H.~Zhang, H.~Liu, J.~Xiao, J.~Dong \emph{et~al.}, ``Mimo: Unlocking the reasoning potential of language model--from pretraining to posttraining,'' \emph{arXiv preprint arXiv:2505.07608}, 2025.

\bibitem{abdin2024phi}
M.~Abdin, J.~Aneja, H.~Behl, S.~Bubeck, R.~Eldan, S.~Gunasekar, M.~Harrison, R.~J. Hewett, M.~Javaheripi, P.~Kauffmann \emph{et~al.}, ``Phi-4 technical report,'' \emph{arXiv preprint arXiv:2412.08905}, 2024.

\bibitem{Mistral}
\BIBentryALTinterwordspacing
mistralai, ``Mistral-small-3.1-24b-instruct-2503,'' 2025. [Online]. Available: \url{https://huggingface.co/mistralai/Mistral-Small-3.1-24B-Instruct-2503}
\BIBentrySTDinterwordspacing

\bibitem{yang2025r4b}
Q.~Yang, B.~Ni, S.~Xiang, H.~Hu, H.~Peng, and J.~Jiang, ``R-4b: Incentivizing general-purpose auto-thinking capability in mllms via bi-mode annealing and reinforce learning,'' \emph{arXiv preprint arXiv:2508.21113}, 2025.

\bibitem{lu2025ovis2}
S.~Lu, Y.~Li, Y.~Xia, Y.~Hu, S.~Zhao, Y.~Ma, Z.~Wei, Y.~Li, L.~Duan, J.~Zhao \emph{et~al.}, ``Ovis2. 5 technical report,'' \emph{arXiv preprint arXiv:2508.11737}, 2025.

\bibitem{li2023llavamed}
C.~Li, C.~Wong, S.~Zhang, N.~Usuyama, H.~Liu, J.~Yang, T.~Naumann, H.~Poon, and J.~Gao, ``Llava-med: Training a large language-and-vision assistant for biomedicine in one day,'' \emph{Advances in Neural Information Processing Systems}, vol.~36, pp. 28\,541--28\,564, 2023.

\bibitem{chen2024huatuogpt}
J.~Chen, C.~Gui, R.~Ouyang, A.~Gao, S.~Chen, G.~H. Chen, X.~Wang, R.~Zhang, Z.~Cai, K.~Ji \emph{et~al.}, ``Huatuogpt-vision, towards injecting medical visual knowledge into multimodal llms at scale,'' \emph{arXiv preprint arXiv:2406.19280}, 2024.

\bibitem{xu2025lingshu}
W.~Xu, H.~P. Chan, L.~Li, M.~Aljunied, R.~Yuan, J.~Wang, C.~Xiao, G.~Chen, C.~Liu, Z.~Li \emph{et~al.}, ``Lingshu: A generalist foundation model for unified multimodal medical understanding and reasoning,'' \emph{arXiv preprint arXiv:2506.07044}, 2025.

\bibitem{pan2025medvlm}
J.~Pan, C.~Liu, J.~Wu, F.~Liu, J.~Zhu, H.~B. Li, C.~Chen, C.~Ouyang, and D.~Rueckert, ``Medvlm-r1: Incentivizing medical reasoning capability of vision-language models (vlms) via reinforcement learning,'' in \emph{International Conference on Medical Image Computing and Computer-Assisted Intervention}.\hskip 1em plus 0.5em minus 0.4em\relax Springer, 2025, pp. 337--347.

\bibitem{lai2025medr1}
Y.~Lai, J.~Zhong, M.~Li, S.~Zhao, and X.~Yang, ``Med-r1: Reinforcement learning for generalizable medical reasoning in vision-language models,'' \emph{arXiv preprint arXiv:2503.13939}, 2025.

\bibitem{sunchiron}
H.~Sun, Y.~Jiang, W.~Lou, Y.~Zhang, W.~Li, L.~Wang, M.~Liu, L.~Liu, and X.~Wang, ``Chiron-o1: Igniting multimodal large language models towards generalizable medical reasoning via mentor-intern collaborative search,'' in \emph{The Thirty-ninth Annual Conference on Neural Information Processing Systems}.

\bibitem{sellergren2025medgemma}
A.~Sellergren, S.~Kazemzadeh, T.~Jaroensri, A.~Kiraly, M.~Traverse, T.~Kohlberger, S.~Xu, F.~Jamil, C.~Hughes, C.~Lau \emph{et~al.}, ``Medgemma technical report,'' \emph{arXiv preprint arXiv:2507.05201}, 2025.

\bibitem{lin2025healthgpt}
T.~Lin, W.~Zhang, S.~Li, Y.~Yuan, B.~Yu, H.~Li, W.~He, H.~Jiang, M.~Li, X.~Song \emph{et~al.}, ``Healthgpt: A medical large vision-language model for unifying comprehension and generation via heterogeneous knowledge adaptation,'' \emph{arXiv preprint arXiv:2502.09838}, 2025.

\bibitem{tang2024medagents}
X.~Tang, A.~Zou, Z.~Zhang, Z.~Li, Y.~Zhao, X.~Zhang, A.~Cohan, and M.~Gerstein, ``Medagents: Large language models as collaborators for zero-shot medical reasoning,'' in \emph{Findings of the Association for Computational Linguistics: ACL 2024}, 2024, pp. 599--621.

\bibitem{hurst2024gpt4v}
A.~Hurst, A.~Lerer, A.~P. Goucher, A.~Perelman, A.~Ramesh, A.~Clark, A.~Ostrow, A.~Welihinda, A.~Hayes, A.~Radford \emph{et~al.}, ``Gpt-4o system card,'' \emph{arXiv preprint arXiv:2410.21276}, 2024.

\bibitem{Qwen-VL}
J.~Bai, S.~Bai, S.~Yang, S.~Wang, S.~Tan, P.~Wang, J.~Lin, C.~Zhou, and J.~Zhou, ``Qwen-vl: A versatile vision-language model for understanding, localization, text reading, and beyond,'' \emph{arXiv preprint arXiv:2308.12966}, 2023.

\bibitem{lu2024deepseek}
H.~Lu, W.~Liu, B.~Zhang, B.~Wang, K.~Dong, B.~Liu, J.~Sun, T.~Ren, Z.~Li, H.~Yang \emph{et~al.}, ``Deepseek-vl: towards real-world vision-language understanding,'' \emph{arXiv preprint arXiv:2403.05525}, 2024.

\bibitem{glm2024chatglm}
T.~GLM, A.~Zeng, B.~Xu, B.~Wang, C.~Zhang, D.~Yin, D.~Rojas, G.~Feng, H.~Zhao, H.~Lai, H.~Yu, H.~Wang, J.~Sun, J.~Zhang, J.~Cheng, J.~Gui, J.~Tang, J.~Zhang, J.~Li, L.~Zhao, L.~Wu, L.~Zhong, M.~Liu, M.~Huang, P.~Zhang, Q.~Zheng, R.~Lu, S.~Duan, S.~Zhang, S.~Cao, S.~Yang, W.~L. Tam, W.~Zhao, X.~Liu, X.~Xia, X.~Zhang, X.~Gu, X.~Lv, X.~Liu, X.~Liu, X.~Yang, X.~Song, X.~Zhang, Y.~An, Y.~Xu, Y.~Niu, Y.~Yang, Y.~Li, Y.~Bai, Y.~Dong, Z.~Qi, Z.~Wang, Z.~Yang, Z.~Du, Z.~Hou, and Z.~Wang, ``Chatglm: A family of large language models from glm-130b to glm-4 all tools,'' 2024.

\bibitem{he2024meddr}
S.~He, Y.~Nie, Z.~Chen, Z.~Cai, H.~Wang, S.~Yang, and H.~Chen, ``Meddr: Diagnosis-guided bootstrapping for large-scale medical vision-language learning,'' \emph{arXiv preprint arXiv:2404.15127}, vol.~1, no.~3, p.~6, 2024.

\bibitem{gpt5_4}
\BIBentryALTinterwordspacing
OpenAI, ``Gpt-5.4,'' 2026. [Online]. Available: \url{https://openai.com/zh-Hans-CN/index/introducing-gpt-5-4/}
\BIBentrySTDinterwordspacing

\bibitem{team2026kimi}
K.~Team, T.~Bai, Y.~Bai, Y.~Bao, S.~Cai, Y.~Cao, Y.~Charles, H.~Che, C.~Chen, G.~Chen \emph{et~al.}, ``Kimi k2. 5: Visual agentic intelligence,'' \emph{arXiv preprint arXiv:2602.02276}, 2026.

\bibitem{minimax27}
\BIBentryALTinterwordspacing
MiniMax, ``Minimax m2.7,'' 2026. [Online]. Available: \url{https://www.minimaxi.com/models/text/m27}
\BIBentrySTDinterwordspacing

\bibitem{qwen3.5}
\BIBentryALTinterwordspacing
Q.~Team, ``Qwen3.5,'' 2026. [Online]. Available: \url{https://qwen.ai/blog?id=qwen3.5}
\BIBentrySTDinterwordspacing

\bibitem{zhang2023huatuogpt}
H.~Zhang, J.~Chen, F.~Jiang, F.~Yu, Z.~Chen, G.~Chen, J.~Li, X.~Wu, Z.~Zhiyi, Q.~Xiao \emph{et~al.}, ``Huatuogpt, towards taming language model to be a doctor,'' in \emph{Findings of the association for computational linguistics: EMNLP 2023}, 2023, pp. 10\,859--10\,885.

\bibitem{jiang2025hulu}
S.~Jiang, Y.~Wang, S.~Song, T.~Hu, C.~Zhou, B.~Pu, Y.~Zhang, Z.~Yang, Y.~Feng, J.~T. Zhou \emph{et~al.}, ``Hulu-med: A transparent generalist model towards holistic medical vision-language understanding,'' \emph{arXiv preprint arXiv:2510.08668}, 2025.

\end{thebibliography}

% \bibitem{b1} G. O. Young, ``Synthetic structure of industrial plastics,'' in \emph{Plastics,} 2\textsuperscript{nd} ed., vol. 3, J. Peters, Ed. New York, NY, USA: McGraw-Hill, 1964, pp. 15--64.

% \end{thebibliography}

\end{document}